\documentclass[preprint,11pt,authoryear]{elsarticle}

\usepackage{amsmath}
\usepackage{fontenc}
\usepackage{subfig}
\usepackage{bm}
\usepackage{mathtools}
\DeclarePairedDelimiter{\norm}{\lVert}{\rVert}
\usepackage{graphicx}
\usepackage{stackengine}
\usepackage{wrapfig}
\usepackage{amssymb}
\usepackage[dvipsnames]{xcolor}
\usepackage{xcolor, soul}
\def\bfm#1{{\bf #1}}

\def\bfs#1{\mbox{\boldmath{$ #1 $}}}

\newcommand\rd{d}

\definecolor{fgwhite}{rgb}{1,1,1}     % white color
\definecolor{fgred}{rgb}{0.8,0,0}     % red color
\definecolor{fgorange}{rgb}{0.93,0.53,0.18}     % orange color
\definecolor{fgpurple}{rgb}{0.55,0.1,0.6}     % purple color
\definecolor{fggreen}{rgb}{0,0.5,0}     % green color
\definecolor{bggreen}{rgb}{0.8,1,0.8}     % pale green color
\definecolor{fgblue}{rgb}{0,0,0.7}     % blue color
\definecolor{bgblue}{rgb}{0.9,0.9,1}     % pale blue color
\definecolor{fgclay}{rgb}{0.51,0.25,0.04}     % clay color

\sethlcolor{green}

\setcitestyle{square}
\usepackage{lineno,hyperref}
\journal{CMAME}

\biboptions{authoryear}

\begin{document}

\begin{frontmatter}

\title{Probabilistic Deep Learning for Real-Time Large Deformation Simulations}
% \tnotetext[mytitlenote]{Fully documented templates are available in the elsarticle package on \href{http://www.ctan.org/tex-archive/macros/latex/contrib/elsarticle}{CTAN}.}

\author[mymainaddress]{Saurabh Deshpande}

\author[mymainaddress,mysecondaryaddress]{Jakub Lengiewicz}

\author[mymainaddress]{St\'ephane P.A. Bordas}
% \cortext[mycorrespondingauthor]{Corresponding author}
% \ead{stephane.bordas@alum.northwestern.edu}

\address[mymainaddress]{Department of Engineering; Faculty of Science, Technology and Medicine; University of Luxembourg}
\address[mysecondaryaddress]{Institute of Fundamental Technological Research, Polish Academy of Sciences}

\begin{abstract}
For many novel applications, such as patient-specific computer-aided surgery, conventional solution techniques of the underlying nonlinear problems are usually computationally too expensive and are lacking information about how certain can we be about their predictions. In the present work, we propose a highly efficient deep-learning surrogate framework that is able to accurately predict the response of bodies undergoing large deformations in real-time. The surrogate model has a convolutional neural network architecture, called U-Net, which is trained with force-displacement data obtained with the finite element method. We propose deterministic and probabilistic versions of the framework. The probabilistic framework utilizes the Variational Bayes Inference approach and is able to capture all the uncertainties present in the data as well as in the deep-learning model. Based on several benchmark examples, we show the predictive capabilities of the framework and discuss its possible limitations.

\end{abstract}

\begin{keyword}
Convolutional Neural Network, Bayesian Inference, Bayesian Deep Learning, Large deformations, Finite Element Method, Real-Time Simulations 
\end{keyword}

\end{frontmatter}

% \linenumbers

\section{Introduction}

Reliable and computationally efficient models are crucial in the design, optimization, or control for various application domains, including aerospace engineering, robotics, or bio-medicine. For instance the increasing interest in biomedical simulations~\citep{ayache}, \citep{delingette}, \citep{COURTECUISSE2014394}, \citep{real_ex1}, \citep{bui} may require having real-time responses. Finding convenient trade-offs between the accuracy and response time of such computational models is currently an active area of research in the context of digital twins, and is also one of the motivations for the research presented in this work.

When accuracy is important, the most general and widely used methodology in engineering for solving boundary value problems is the finite element method (FEM)~\citep{zienkiewicz1991finite}. This accuracy, especially when it comes to highly non-linear or history-dependent problems, may require a significant computational effort. The advancements in hardware development and software optimization enabled to some extent speeding up FEM computations, which involves specialized solution strategies to take advantage of high-performance computing architectures. A notable example is the class of Finite Element Tearing and Interconnecting (FETI) methods~\citep{FETI}. In these methods, the global domain is partitioned into a set of disconnected sub-domains, which are computed in parallel on different processors/nodes. However, in many applications, it is not possible to meet real-time responses on the hardware available for industrial consumers. This is due to a limited number of available cores and a significant communication burden that deteriorates the overall time performance of such solution strategies.

There are various specialized FEM-based approaches to cut down the solution time at the cost of sacrificing the accuracy, see, e.g., ~\citep{fastFEM}. An important class of such approaches is the model order reduction (MOR) methods, with Proper Orthogonal Decomposition (POD) being one of the notable examples. The general concept of POD, e.g., applied to a discretized FEM formulation, is to ﬁnd a low dimension subspace in order to approximate the full space at an acceptable loss of accuracy. This potentially enables controlling the trade-off between accuracy and computation time. POD was adapted to work within the large-deformation regime, see, e.g., \citep{KERFRIDEN2011850}, which was for instance applied to simulate and control soft robotic arms~\citep{PODSoft} or to reduce computational costs in nonlinear fracture mechanics problems~\citep{MOR1}. However, the efficiency of POD deteriorates in high non-linear regimes since it relies on a linear combination of few basis vectors and thus oversimplifies the model \citep{BHATTACHARJEE2016635}. Proper Generalised Decomposition (PGD) is another MOR technique, in which the solution of the complete problem is computed as a finite sum of separable functions. The compact solution, though not optimal, in general, provides a very light format to store the solution in the form of a meta-model, thereby speeding up the solution times. \citep{PGD} implemented PGD based approach for computationally efficient simulations of hyper-elastic responses. However, the accuracy of PGD methods decreases when the separation of variables assumption cannot respect the problem to solve \citep{PGD_drawback}. 

Importantly for the present work, we distinguish yet another family of FEM-based approaches, in which the expected speedups and approximation capabilities originate from underlying Deep Neural Networks (DNNs) with Deep Learning (DL) techniques used to train these networks. Generally, DL approaches make an important part of machine learning techniques and have allowed to solving highly complex problems that had eluded scientists for decades. In particular, DL-based methods have also been developed to efficiently solve problems in engineering~\citep{DeepBook}. One of the popular approaches utilises the idea of the so-called Physics Informed Neural Networks (PINNs), in which both the data (either synthetic or experimental) and the assumed governing Partial Differential Equations (PDEs) are incorporated in the training phase; for early traces of these see, e.g., ~\citep{earlyPINN2}, \citep{PINN3}, \citep{earlyPINN}, and for a recent study see, e.g., \citep{PINN},\citep{SAMANIEGO2020112790}. One of the benefits of this approach is that possibly much less data is needed for training, which can be an important factor for many data-driven applications. Note, however, that even if the physics is not explicitly enforced in the training phase (non-PINN case), it can still be recovered in the trained model by implicitly following a large amount of training data~(synthetic or experimental). In the case when all training data are synthetically provided from FEM simulations, see, e.g., \citep{LORENTE2017342}, \citep{cotin}\citep{vasilis}, we will refer to it as the \emph{direct FEM-based approach}. 

In this work, we propose a framework that falls into the above-mentioned class of direct FEM-based DNN approaches. The framework is based on a particular DNN architecture---the U-Net architecture~\citep{unet_original}---which in turn can be viewed as a type of Convolutional Neural Networks (CNNs); further explanations are provided later in this work. Originally, the U-Net architecture has been developed for the purpose of biomedical image segmentation, however, it turned out to be also suitable for other applications. In particular, the present work is inspired by the recent results of~\citep{cotin}, in which the authors demonstrate quite accurate real-time non-linear force-displacement predictions done by U-Nets trained on FEM-based data. It has been noticed, see~\citep{he2019mgnet}, \citep{unet_nature}, that this good accuracy is not accidental but can be possibly linked to the strong resemblance of U-Net architectures and multi grid solution schemes~\citep{multigrid_book}. Such a point of view makes the U-Net approach less of a brute-force black-box approximation and more of a suited solution scheme, which makes this line of research very promising.

An equally important aspect that is studied in the present work is the capabilities of DNNs to quantify uncertainties. This is motivated by the fact that in many real-life applications, such as surgical simulations~\citep{bui} or autonomous driving~\citep{BDL_auto}, it is crucial to produce reliable uncertainty estimates in addition to the predictions. Otherwise, model predictions can lead to harmful consequences in these critical tasks. Deterministic neural networks are usually certain about their predictions, and this overconfidence is especially evident when facing data far from the training set. Generally, uncertainties can be categorised as those associated with the misfit of neural network models (epistemic uncertainties) and those that refer to the noise in training data (aleatoric uncertainties)\citep{survey_uncertainty}\citep{alex_uncertainty}. Uncertainties can either fall within or outside the training data region (interpolated and extrapolated regions, respectively). Within the data region, the prediction uncertainties are caused both by noisy data and by the model misfit, which can be captured and quantified in many ways, the Maximum Likelihood Estimation (MLE) method is one of the most straightforward~\citep{10.5555/3295222.3295309} to do so. However, for the extrapolated region we have no data support, therefore, no direct quantification can be done there. What we can only reasonably assume is that the uncertainty should generally increase when moving away from the data region. To achieve this, in this work, we will extend the idea proposed in \citep{bayesbackprop}\citep{Gal2016UncertaintyID}\citep{duerr2020probabilistic} which relies on converting a neural network to its stochastic counterpart by replacing discrete parameters with probability distributions and using a special training technique that is based on Bayesian Inference.   

Bayesian approaches have been already considered in the context of FEM models; for instance in \citep{paul} uncertainties are quantified for hyperelastic soft tissues by incorporating stochastic parameters in the FEM model, while in \citep{hus}\citep{zeraatpisheh} a thorough tutorial on using Bayesian Inference to solid mechanics problems is provided. In the present work, however, we focus on Bayesian Inference and related Bayesian Neural Networks (BNNs). Here, in the context of deep networks with millions of parameters, the application of widely used Markov Chain Monte Carlo~(MCMC) type of methods would be computationally intractable. For that reason, in this work we use one of the well-known methods for approximate Bayesian inference---Variational Inference (VI) \citep{VI_graves}---in which an approximate distribution is used instead of the true Bayesian posterior over model parameters. 

To sum up, the scope of the present work is to study the applicability of U-Net deep learning architectures to performing real-time predictions for large-deformation problems with uncertainties, where synthetic training data is provided by FEM simulations. 
The organisation of the paper is the following. In Section~\ref{sec: Methodology} we present the general methodology applied to a deterministic version of U-Net. In Section~\ref{sec: BDL} we introduce the extension of the framework to the Variational Bayesian Inference case. Then, in Section~\ref{sec: Results}, an extensive study of the proposed framework is performed, which is based on several 2D and 3D benchmark examples. The conclusions and future research directions are outlined in Section~\ref{sec: Conclusion}.

%%%%%%%%%%%%%%%%%%%%%%%
%%%%%%%%%%%%%%%%%%%%%%%
\section{General FEM-based U-Net Methodology}
\label{sec: Methodology}

The proposed approach can be divided into two main phases. The first phase involves finite element simulations to prepare necessary datasets. The second phase consists of building and training deterministic/probabilistic U-Net deep neural networks, using the training datasets generated in the first phase. The trained U-Nets are then used as surrogate models. 

\begin{figure}[h]
     \centering
     \includegraphics[scale=0.6]{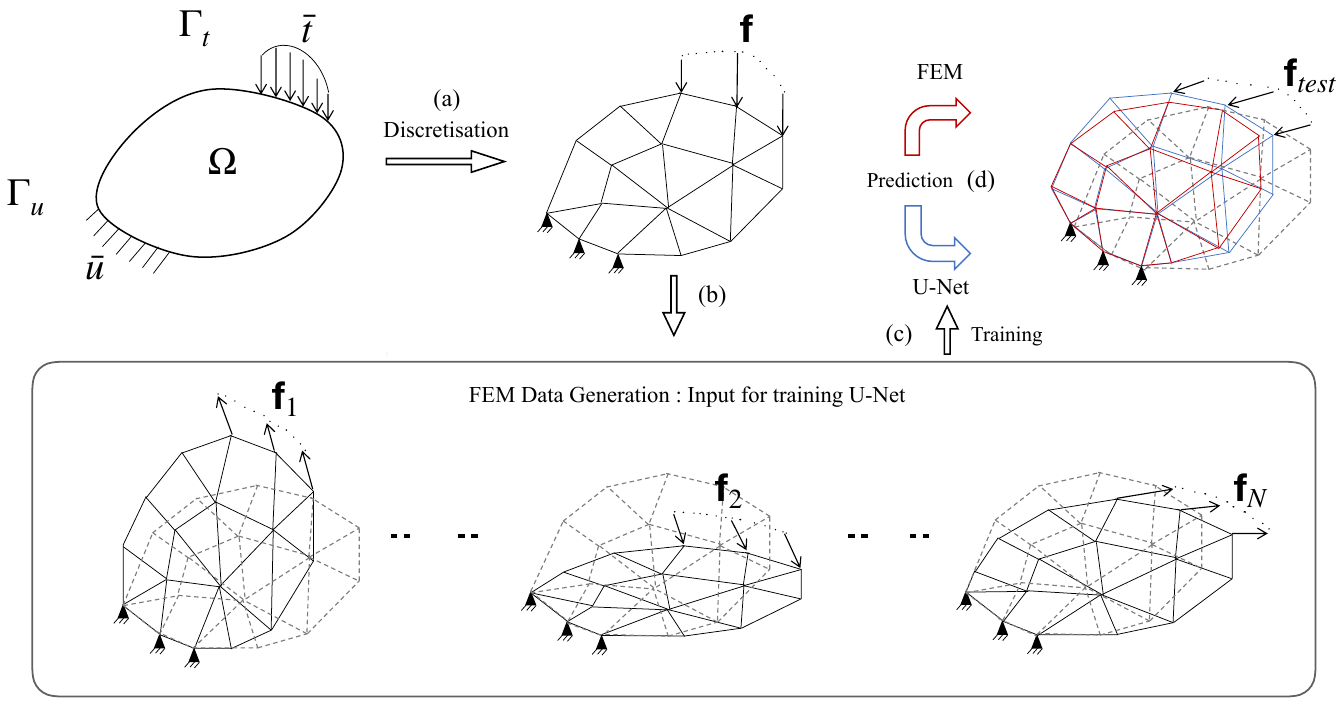}
     \caption{Schematic of the framework (a) Continuum problem is discretized by FEM mesh (b) Training/testing examples are generated by applying random point forces on Neumann boundary. (c) The U-Net is trained on the generated dataset (d) Trained U-Net predicts the deformation for a test force (blue mesh). FEM solution (red) is used for cross-validation. Gray dashed meshes indicate undeformed configurations.}
     \label{schematic}
\end{figure}

%%%%%%%%%%%%%%%%%%%%%%%
\subsection{FEM-Based Deep Learning Approach}
\label{sec:fem_based_framework_intro}

As a problem to be solved, we consider the boundary value problem of a hyperelastic solid, with a constant prescribed Dirichlet BC. Large deformations exhibit non-linear stress-strain behavior when applied with external forces, hence one needs to consider hyper-elastic constitutive laws for simulating such systems. 

Consider a boundary value problem in continuum mechanics in the domain $\Omega$, Dirichlet and Neumann boundary conditions are applied on $\Gamma_D$, $\Gamma_N$ respectively. Neglecting the body forces, the virtual work principle for nonlinear elastostatic equation reads

  \begin{equation}
    \label{eq:VWP}
          \int_{\Omega} \bfm{P}(\bfs{u}) \cdot \nabla\delta\bfs{u}\, \rd V -
      \int_{\Gamma_t} \bfs{\bar{t}} \cdot \delta\bfs{u}\, \rd S  = 0 \quad\quad\forall\delta\bfs{u},
  \end{equation}
where $\bfs{u}$ and $\delta\bfs{u}$ belong to appropriate functional spaces, $\bfs{u}=\bar{\bfs{u}}$ and $\delta\bfs{u}=\bfm{0}$ on $\Gamma_u$, and $\bfm{P}(\bfs{u})$ is the first Piola-Kirchhoff stress tensor. The required constitutive relationship will be defined through the (hyper-)elastic strain energy potential $W(\bfs{F})$ as
  \begin{equation}
    \label{eq:const}
   \bfm{P}(\bfs{F}) = \frac{\partial W(\bfs{F})}{\partial \bfs{F}}
  \end{equation}
where $\bfs{F}=\bfs{I} + \nabla \bfs{u}$ is the deformation gradient tensor. (The particular form of $W$, used in this work, will be presented in Section~\ref{data_generation_FEM}.) 

After standard FE discretization, the problem expressed by Eq.~(\ref{eq:VWP}) will take the form of system of non-linear equations 
  \begin{equation}
    \label{eq:res}
        \mathbf{R}(\mathbf{u})=\mathbf{f}_{\text{int}}(\bfm{u})-\mathbf{f}_{\text{ext}}=\bfm{0}, 
  \end{equation}
which expresses the balance between external and internal nodal forces. By solving the system of equations~(\ref{eq:res}) (e.g., with the Newton-Raphson method) for a given external force vector $\mathbf{f}_{\text{ext}}=\bfm{f}$ we obtain a solution in the form of nodal displacements $\mathbf{u}$.

\par

External forces can be applied to a selected region on the surface described by $\Gamma_t$. For the current framework, we consider a single FE discretization of a given domain $\Omega$. As described in Figure~\ref{schematic}, we apply a prescribed family of load distributions, given by vectors of force $\bfm{f}_i$ on the nodes present in $\Gamma_t$ to generate nodal displacements $\bfm{u}_i$. This creates a dataset $\mathcal{D} = {(\bfm{f}_i,\bfm{u}_i)}_{i=1}^{N}$ of corresponding pairs of nodal force and displacement vectors, which is then used as input to train DNNs. Thus the input of the neural network is a vector of nodal forces, and the predicted output is a vector of nodal displacements.

%%%%%%%%%%%%%%%%%%%%%%%
\subsection{U-Net deep neural network architecture}\label{sec:unet_architecture}

\begin{figure}[h]
     \centering
     \includegraphics[width=\textwidth]{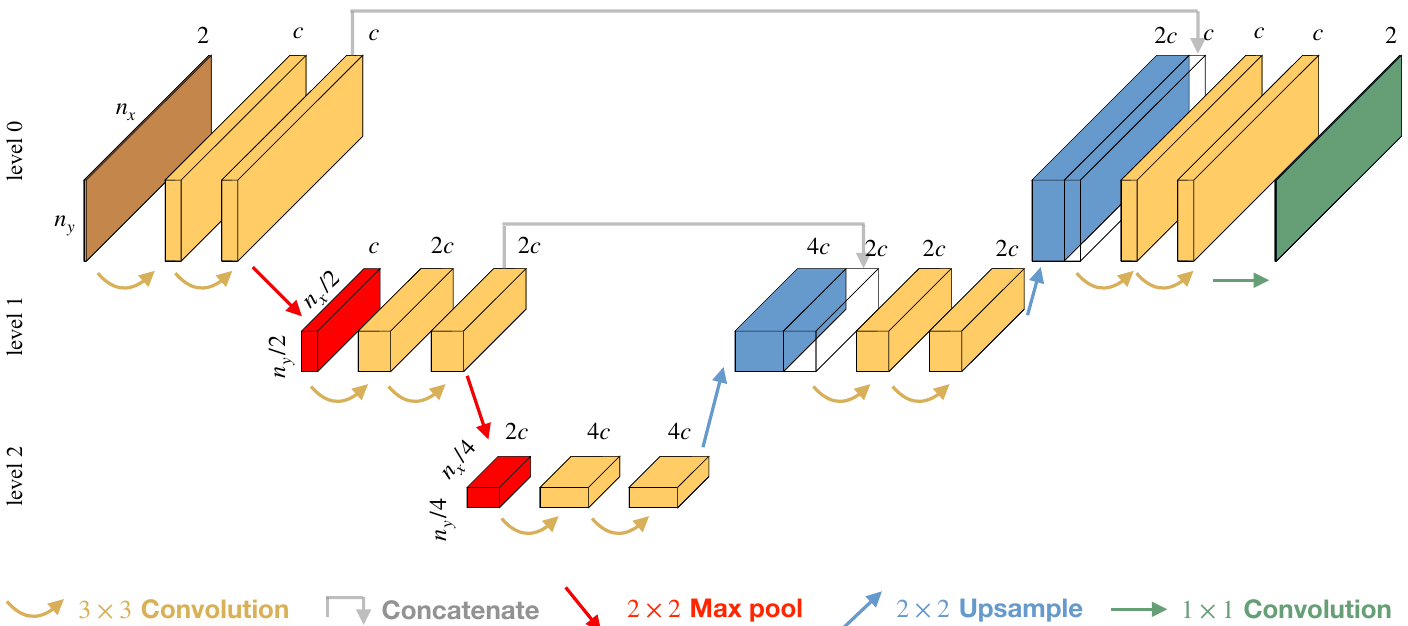}
     \caption{A schematic of exemplary U-Net architecture for 2D domains, ($n_{\text{x}}, n_{\text{y}}$) stand for number of nodes in x, y direction of 2D domain. Boxes indicate U-Net layers (colors indicate different types of layers), first step contains '$c$' channels.}
     \label{fig:CNN-2D}
\end{figure}

As motivated in the introduction, we use a specific family of CNN, U-Nets~\citep{unet_original}. They owe their name to a specific U-shape of the architecture diagram, e.g., see Figure~\ref{fig:CNN-2D}, which is an effect of applying cascades of max pooling operations, followed by cascades of upsampling operations.

At its input, the layer $\bfs{d}^0$, the U-Net network, $\mathcal{U}$, accepts the vector of external forces, $\bfm{f}$, in the original mesh format $n_x \times n_y \times n_z \times 3$, where $n_x$, $n_y$, $n_z$ are the dimensions of the structured 3D mesh (for 2D problems the format is $n_x \times n_y \times 2$). The output displacements, in the layer $\bfs{d}^L$, are in the analogous mesh format.
%, which is then transformed to the displacement vector $\bfm{u}$. 
The model parameters $\bfs{\theta}_{\text{det}}$ are all weights (kernel $\bfs{k}$ and biases $\bfs{b}$) of the neural network (see below for details).

For the sake of clarity, below we will introduce the idea for a 2D case only, which can be straightforwardly transformed into a 3D case. For 2D problems, we use architecture as described in Figure~\ref{fig:CNN-2D}. To a given input mesh, we first add double zero padding in each spatial direction (this is done to avoid the loss of information of corner nodes). There are two layers in each encoding and decoding phase. To the padded input, we apply two convolutions with batch normalisation followed by rectified linear unit activation~(ReLU), see Eq.~(\ref{Eq: 3x3 Conv}). In the encoding phase, at each step, $2\times2$ Max Pooling layers are applied which decrease spatial dimensions by half, and we multiply the number of channels by 2 (c=128 for the first level). In decoding steps, $2\times2$ Up-sampling layers are applied which increases the spatial dimension by two and the number of channels is halved. At each level, encoding and decoding outputs are concatenated together. In the end, $1\times1$ convolution is applied with linear activation to get the output. 

\begin{figure}[h]
     \centering
     \includegraphics[width=0.4\textwidth]{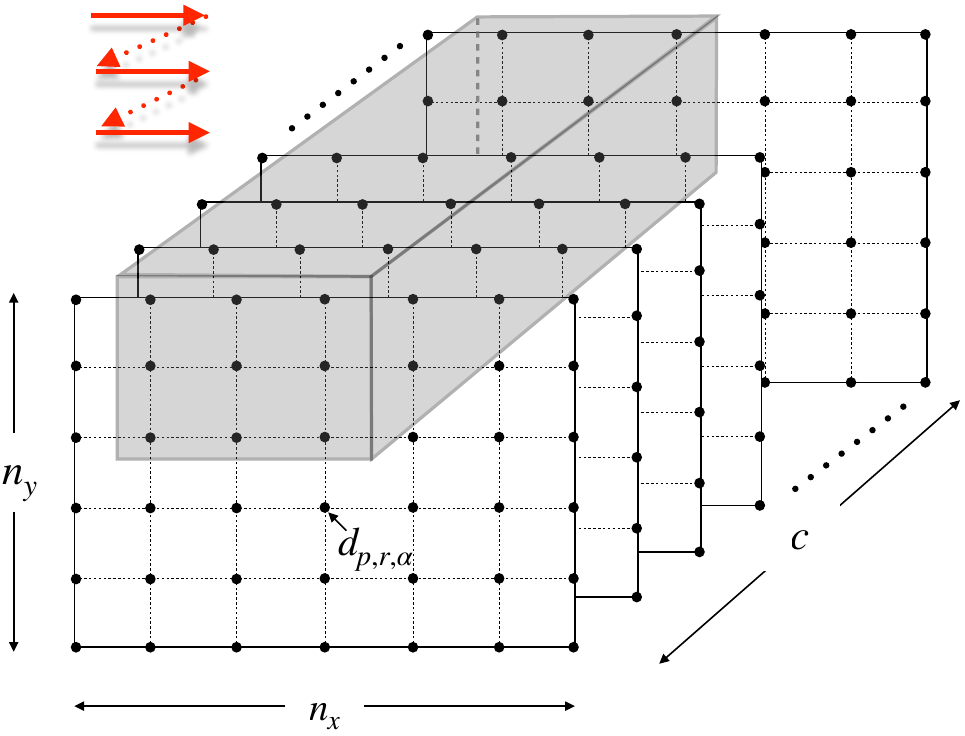}
     \caption{$x$ and $y$ dofs are stored in different channels, $3\times3$ convolutional filter (gray) acts locally along the channel direction and it slides along with the step of 1 in both horizontal and vertical directions (red).}
     \label{fig:filter_application}
\end{figure}

In order to better understand the idea of convolution operator, the $3\times 3 \times c$ operator in 2D on the first U-Net level can be imagined as a filter window that is applied to a $n_x \times n_y \times c$ mesh. For a two-dimensional domain, the input tensor is of the dimension $n_x \times n_y \times 2$, which is identical to the FEM mesh. Here $2$ stands for the number of channels of input convolutions, $n_x, n_y$ stands for the number of nodes in the $x$, $y$ direction, respectively. To best leverage the CNNs, we keep $x$ \& $y$-dofs in separate channels. We operate a convolutional filter on a local region and then is slid along spatial directions $x,y$ with a stride of 1 as illustrated in Figure~\ref{fig:filter_application}.

An example of a non-restrictive convolution operation (in 2D), between subsequent U-Net layers $l$ and $l+1$, for the filter size $3\times{}3\times{}c^{n}$ reads

\begin{equation}
    \label{Eq: 3x3 Conv}
    d^{l+1}_{p,r,\beta}=\mathcal{A}\left(b^{l+1}_{\beta}+\sum_{i=1,}^{3}\sum_{j=1,}^{3}\sum_{\alpha=1}^{c^{l}} d^{l}_{p+i-2,r+j-2,\alpha} k^{l+1}_{i,j,\alpha,\beta}\right),
\end{equation}

where $d^{l+1}_{p,r,\beta}$ are neural network nodes at layer $l+1$, the weights $k^{l+1}_{i,j,\alpha,\beta}$ are parameters of the convolution operator, the weights $b^{l+1}_{\beta}$ are biases at a layer $l+1$, and $\mathcal{A}(\cdot)$ is an activation function (ReLU). Indices $i,j$ stand for the components of the covolutional filter ($3\times3$ in our case) and the indices $p,r$ are related to nodes in a 2D grid of the output layers. They directly refer to the underlying structured FEM mesh. In our case, we add zero pad in each dimension of input before applying the convolution, to ensure the same size of the input and output. Indices $1\leq{}\alpha\leq{c^{l}}$ and $1\leq{}\beta\leq{c^{l+1}}$ represent the channel number. Note that the number of channels in subsequent layers need not be equal, i.e., in general $c^{l}\neq{}c^{l+1}$. Note also that in the first and in the last layer, the number of channels correspond to the spatial dimension of the problem (2D or 3D). 

Max-pooling operation is responsible for reducing the spatial dimensions of its input, channel dimensions are unaffected by it. It reads as follows 

\begin{equation}
    d^{l+1}_{p,r,\alpha}= \max\limits_{\substack{2p-1\leqslant i \leqslant 2p \\[0.2em]{2r-1\leqslant j \leqslant 2r}}} d_{i,j,\alpha}^{l} 
\end{equation}

Upsampling operator can be seen as the reverse of max pooling, it increases the spatial dimensions of the input without affecting the channel dimension. As showed in Figure~\ref{fig:CNN-2D}, in the decoder phase, outputs of up-sampling are concatenated with respective layers from the encoder phase of the U-Net (in case of symmetric U-Nets, the $l$-th layer is concatenated with the $(L-l-2)$-th layer, where $L$ is the index of output layer in a U-Net). The up-sampling with concatenation read 

\begin{equation}
     d_{p,r,\alpha}^{l+1}= 
\begin{cases}
    d_{{\lfloor}p/2{\rfloor}+1,{\lfloor}p/2{\rfloor}+1,\alpha}^{l}~~, &  1\leqslant \alpha \leqslant c^{l} \\[1em]
    d_{p,r,(\alpha-c^{l})}^{L-l-2}~~              &  c^{l}+1\leqslant \alpha \leqslant c^{l}+c^{L-l-2}
\end{cases}
\label{eq:upsampling}
\end{equation}

The final $1 \times 1$ convolution operation reads 

\begin{equation}
    d^{L}_{p,r,\beta}=b^{L}_{\beta}+ \sum_{\alpha=1}^{c^{L-1}}  d^{L-1}_{p,r,\alpha} k^{L}_{\alpha,\beta}, \quad\quad \beta=1,2.
    \label{eq:1x1_conv}
\end{equation}

For the 3D version of U-Nets, the operations given by Equations~(\ref{Eq: 3x3 Conv})-(\ref{eq:1x1_conv}) are straightforwardly extended by one additional dimension. This results in adding one index to nodes' and biases' specifications, $d$ and $b$, respectively, and two indexes to $3\times{}3$ convolution weights, $k$. For 2D/3D cases, trainable parameters of the deterministic U-Net are

\begin{equation}
    \bm{\theta}_{\text{det}}=\bigcup_{l=1}^L \{\bfm{k}^l,\bfm{b}^l\}.
    \label{eq:det_params}
\end{equation}
$\mathcal{U}(\bfm{f},\bm{\theta}^{\text{det}})$ is defined recursively (the forward propagation), starting from the input layer $\bfs{d}^0=\bfm{f}$, then subsequently applying appropriate transformations given by one of Eqs.~(\ref{Eq: 3x3 Conv})-(\ref{eq:upsampling}), and finally applying the transformation given by Eq.~(\ref{eq:1x1_conv}), see also Figure~\ref{fig:CNN-2D}. Finally the prediction of the deterministic U-Net is 

\begin{equation}
    \mathcal{U}(\bfm{f},\bm{\theta}_{\text{det}})=\bfm{d}^L
\end{equation}

For a given training dataset $\mathcal{D}=\{(\bfm{f}_1,\bfm{u}_1),...,(\bfm{f}_N,\bfm{u}_N)\}$, the deterministic U-Net is trained by minimizing the following mean squared error loss function
\begin{equation}
      \mathcal{L}_{\text{det}}(\mathcal{D},\bm{\theta}_{\text{det}}) = \frac{1}{N}\sum_{i=1}^{N} \norm{\mathcal{U}(\bfm{f}_i,\bm{\theta}_{\text{det}})-\bfm{u}_i}_2^{2} \label{eq:lossDeterm}
\end{equation}
which gives the optimal parameters  
\begin{equation}
      \bm{\theta}^{*}_{\text{det}} = \text{arg }\underset{\bm{\theta}_{\text{det}}}{\text{min}}~\mathcal{L}_{\text{det}}(\mathcal{D},\bm{\theta}_{\text{det}}). \label{eq:minimisationLossDeterm}
\end{equation}
A particular training strategy, used in this work, is introduced in Section~\ref{sec:training_UNets}.

Remark: The current framework (as well as all other neural-network based approaches mentioned in the introduction) requires retraining a network when changing the FE discretization. Recently proposed operator-based learning approaches, called neural operators~\citep{DeepONet}\citep{li2021fourier}\citep{UFNO}, promise to overcome this disadvantage.

\section{Probabilistic U-Net framework}\label{BDL}\label{sec: BDL}
There are various sources of uncertainties linked to engineering systems. These can be broadly categorised as noises in the observation (aleatoric uncertainty or data uncertainty) and uncertainty in the assumption of our model (epistemic uncertainty or model uncertainty) \citep{alex_uncertainty}. Aleatoric uncertainty is inherent to the data and it can't be reduced, whereas epistemic uncertainty can be reduced by providing more training data. An important part of epistemic uncertainty is being able to tell that the more data we have, the more certain we are about the predictions. This uncertainty is expected to be high while doing predictions on inputs away from the training region. Deterministic U-Nets explained in the Section~\ref{sec:unet_architecture} fail to account for these uncertainties. In order to capture these effects, in this work, we propose the Bayesian approach, which is introduced in this section.

\subsection{Variational Bayesian Inference}

Bayesian methods provide an approach to quantify the uncertainty of prediction in deep neural networks. To do so, in this framework, we replace the deterministic parameters with probability distributions \citep{bayesbackprop}. Originally this idea was only used to prevent overfitting, but was observed to also increase the variability of outputs in the extrapolated region. In order to suitably control the level of introduced perturbations to parameters, we use Bayesian Inference. This gives us a formal theoretical framework, allowing us to apply suitable computational techniques (VI) to efficiently train networks and predict results. The input of a network remains the same, but some of the model parameters become probability distributions (stochastic), and for that reason also the output of the network becomes a distribution over possible outputs. As per the standard Bayesian approach, we specify a prior distribution $P(\bm{w})$ over parameters, we consider  $\mathcal{D}=\{(\bfm{f}_1,\bfm{u}_1),...,(\bfm{f}_N,\bfm{u}_N)\}$ as the given training dataset, then for a new vector $\bfm{f}^\text{test}$, prediction $\bfm{u}^\text{test}$ is given by 

\begin{equation}
    P(\bfm{u}^{\text{test}}|\bfm{f}^{\text{test}},\mathcal{D}) = \int P(\bfm{u}^{\text{test}}|\bfm{f}^{\text{test}},\bm{w})P(\bm{w}|\mathcal{D}) \text{d}\bm{w}
    \label{prediction}
\end{equation}
The Bayesian inference involves the calculation of true parameter posterior $P(\bm{w}|\mathcal{D})$ conditioned over the training data. As mentioned in the introduction, variational inference (VI) is used to approximate true posterior densities in bayesian neural networks~\citep{VI_graves}, i.e. true posterior is approximated by a variational posterior $q(\bm{w}|\bm{\theta})$, parametrized by $\bm{\theta}$. Variational learning finds optimal parameters $\bm{\theta^{*}}$ by minimizing the Kullback-Leibler divergence (KL-divergence) between true and variational posteriors, as per the following equation:
\begin{equation}
    \begin{split}
      \bm{\theta}^{*} &=\text{arg } \underset{\bm{\theta}}{\text{min}} ~ \text{KL}\big[q(\bm{w}|\bm{\theta})||P(\bm{w}|\mathcal{D})\big] \\
         &= \text{arg }\underset{\bm{\theta}}{\text{min}}~ \int q(\bm{w}|\bm{\theta}) \log\frac{q(\bm{w}|\bm{\theta})}{P(\bm{w})P(\mathcal{D}|\bm{w})}dw \\
         &= \text{arg }\underset{\bm{\theta}}{\text{min}}~\text{KL}[q(\bm{w}|\bm{\theta})||P(\bm{w})] - \mathbb{E}_{q(\bm{w}|\bm{\theta})}[\log P(\mathcal{D}|\bm{w})].
    \end{split}
    \label{ELBO}
\end{equation}
The resulting cost function is the loss function for training the neural network. It consists of two parts, first is the prior dependent part represented by the KL-divergence term, it can be referred to as model complexity cost; it tells how close approximate posteriors are to priors. And later is the data-dependent part which can be referred to as likelihood cost, it tells how well the network fits the data. Bayesian neural networks with prior distributions are well known to induce regularisation effect \citep{Neal_BDL_book}; in particular, using Gaussian priors is equivalent to weight decay ($L_2$  regularization)~\citep{gaussian_prior}.\par 

During the forward pass, weights are sampled from the variational posterior $q(\bm{w}|\bm{\theta})$.Now, during the backpropagation, the issue is that one cannot get a gradient of the sampled points,  because the sampling operation cannot be differentiated. To avoid this issue, the following reparameterization trick is used \citep{repara}. A sampled weight, $w$, is obtained by sampling a parameter-free distribution (the unit Gaussian), which is then scaled by a standard deviation $\bm{\sigma}$ and shifted by a mean $\bm{\mu}$.  We parameterise the standard deviation point-wise as $\bm{\sigma} = \log (1 + \text{exp}(\bm{\rho}))$ to have $\bm{\sigma}$ always non-negative. Thus the sample is $\bm{w} = \bm{\mu} + \log (1 + \text{exp}(\bm{\rho}))\odot \bm{\epsilon}$, where $\odot$ is point-wise multiplication, $\bm{\epsilon}$ is drawn from $\bm{\mathcal{N}}(\bm{0},\bm{I})$. Hence the variational posterior parameters are $\bm{\theta} = (\bm{\mu},\bm{\rho})$ \citep{bayesbackprop}. In our framework we use Gaussian priors with its parameters being $(\bm{\mu_p},\bm{\sigma_p})$. For the reasons mentioned in Section~\ref{sec: empirical_bayes}, we include prior means ($\bm{\mu_p}$) in training procedure.  

\subsection{Maximum likelihood estimation}

In addition to the Bayesian approach, we also introduce the Maximum Likelihood Estimation (MLE) method---a popular frequentist approach. We do it to compare both methods in their capabilities to quantify uncertainties. Parameters of the MLE model are deterministic, but we take the double number of outputs compared to the deterministic counterpart. They stand for means and non-constant (heteroscedastic) standard deviations~\citep{duerr2020probabilistic}, thus yielding distributions as the outputs. These non-constant standard deviations can only capture the noises in the data, MLE inherently fails to account for uncertainties in the extrapolated region. The loss function for MLE can be recovered from Equation~(\ref{ELBO}) by removing the KL divergence part (since we don't have distributions on parameters of the MLE model), and the MLE model is trained on the Gaussian negative log-likelihood loss:

\begin{equation}
      \bm{\theta}^{*}_{\text{MLE}} = \text{arg } \underset{\bm{\theta}_{\text{MLE}}}{\text{min}} -\log P(\mathcal{D}|\bm{\theta}_{\text{MLE}}).
    \label{MLE}
\end{equation}

\subsection{Trainable priors: Use of Empirical Bayes} \label{sec: empirical_bayes}

Since NN parameters are latent variables of the model, it is very
difficult to make a proper choice of priors. If one sets the priors far
from their true values, then the posterior may be unduly affected by such
choice. To overcome this, we incorporate Empirical Bayes~(EB) approach \citep{Emperical_bayes},  a method that uses the observed data to estimate the prior hyperparameters. In our approach, in the training phase, we update the prior means, keeping the prior standard deviation constant. Hence we minimize the loss function by also considering gradients with respect to the prior means. This treatment enables us to obtain a good fit to the data, while at the same time giving high prediction uncertainties in the region where little or no data is available.

\subsection{Loss functions for probabilistic U-Net} \label{sec: loss_VB}

We modify the deterministic U-Net architectures by replacing their layers with probabilistic layers, as a result, the output of the network is a probability distribution itself. We choose Gaussian distributions to represent priors and approximate posteriors of probabilistic layers. 
For the Bayesian U-Net, we use loss function as given in  Eq.~(\ref{ELBO}). Expectations of the  Eq.~(\ref{ELBO}) are approximated by $\mathcal{M}$ Monte Carlo samples drawn from the approximate posterior $q(\bm{w}|\bm{\theta})$ as referred below
 \vspace{-5mm}

\begin{equation}
    \begin{split}
      \mathcal{L}_{\text{VB}}  &= ~\text{KL}[q(\bm{w}|\bm{\theta})||P(\bm{w})] - \mathbb{E}_{q(\bm{w}|\bm{\theta})}[\log P(\mathcal{D}|\bm{w})]\\
         &\approx ~ \sum_{i=1}^{\mathcal{M}} \log q(\bm{w}^{(i)}|\bm{\theta}) - \log P(\bm{w}^{(i)}) - \log P(\mathcal{D}|\bm{w}^{(i)})
    \end{split}
    \label{ELBO_sampels}
\end{equation}

If we substitute Gaussian probability density functions, the expression in the RHS of Eq.(\ref{ELBO_sampels}) turns out to be as given in Eq.~(\ref{ELBO_expand}). We consider '$\mathcal{G}$' probabilistic parameters (Gaussian distributions) for our Bayesian U-Net, where every distribution is parameterised by its mean and standard deviation values. Since the standard deviations $\bm{\sigma}$ must be positive, we first train the network on untransformed standard deviations $\bm{\rho}$ which are later transformed to $\bm{\sigma}$ through soft-plus function. Also, for the reasons discussed in Section \ref{sec: empirical_bayes}, we involve prior means, $\bm{\mu_p}$, in the training procedure as well.  Hence parameters to be learned in the training procedure are $\bm{\theta}_{\text{VB}}=(\bm{\theta},\bm{\mu_p})$. Finally, the loss function for the Variational Bayes is given as follows

\begin{equation}
    \begin{split}
      \mathcal{L}_{\text{VB}}(\mathcal{D},\bm{\theta}_{\text{VB}})  \approx ~ \sum_{i=1}^{\mathcal{M}}&\left[ \sum_{j=1}^{\mathcal{G}} \left(- \log  \left(\sqrt{2\pi}~\sigma_{j}^{(i)}\right) - \frac{\left(w_{j}^{(i)}-\mu_{j}^{(i)}\right)^{2}}{2\left(\sigma_{j}^{(i)}\right)^2} + \log \left(\sqrt{2\pi}~\sigma_{p}\right) + \frac{\left(w_{j}^{(i)}-(\mu_{p})_{j}^{(i)}\right)^{2}}{2\sigma_{p}^2}\right)\right.  \\
         &- \left.\sum_{k=1}^{N}\sum_{l=1}^{\mathcal{F}}\left( - \log \left(\sqrt{2\pi}~d^{(l)}_{\sigma}(\bfm{f}^{(k)},\bm{w}^{(i)})\right) - \frac{\left(u^{(k)}_l- d^{(l)}_{\mu}(\bfm{f}^{(k)},\bm{w}^{(i)})\right)^{2}}{2\left(d^{(l)}_{\sigma}(\bfm{f}^{(k)},\bm{w}^{(i)})\right)^2}\right)\right]
    \end{split}
    \label{ELBO_expand}
\end{equation}
where
\begin{equation}
    \begin{split}
         d^{(l)}_{\sigma}(\bfm{f}^{(k)},\bm{w}^{(i)}) &= \log(1+\exp({d^{(l)}_{\rho}(\bfm{f}^{(k)},\bm{w}^{(i)})})),
         \\
         w_{j}^{(i)} &= \mu_{j}^{(i)} + \sigma_{j}^{(i)}  \epsilon_{j}^{(i)},\hspace{0.1\textwidth}  \epsilon_{j}^{(i)} \mathcal~ \sim~ {\mathcal{N}}(0,1),\\
         \sigma_{j}^{(i)} &= \log(1+\exp({\rho_{j}^{(i)}})).
    \end{split}
\end{equation}

($\bm{d_{\mu}}(\bfm{f},\bm{w})$, $\bm{d_{\rho}}(\bfm{f},\bm{w})$) are the outputs at the penultimate layer of the Bayesian U-Net, which stand for means and heteroscadetic (non-constant) standard deviations. And the last output layer is a distribution layer with the same parameters. Since ($\bm{d_{\mu}}(\bfm{f},\bm{w})$, $\bm{d_{\rho}}(\bfm{f},\bm{w})$) are variables in themselves, in order to get the prediction one needs to sample over this output distribution. $N,\mathcal{F}$ are total number of training examples and dof per problem respectively. $\sigma_{p}$ stands for the standard deviation of each the prior, which is kept constant in the training procedure. Optimized parameters, $\bm{\theta}^{*}_{\text{VB}} = (\bm{\theta}^{*},\bm{u}^{*}_{\bm{p}}$), for the Variational bayes case are obtained by minimising the above loss function: 

\begin{equation}
    \bm{\theta}_{\text{VB}}^{*} = \text{arg }\underset{\bm{\theta}_{\text{\tiny VB}}}{\text{min}}~\mathcal{L}_{\text{VB}}(\mathcal{D},\bm{\theta}_{\text{\tiny VB}})
    \label{eq: min_VB}
\end{equation}

Once the optimised parameters are computed, we replace the true posterior $P(\bm{w}|\mathcal{D})$ in Eq.~(\ref{prediction}) with the variational posterior $q(\bm{w}|\bm{\theta}^{*})$ to get the predictive distribution: 

\begin{equation}
      P(\bfm{u}|\bfm{f},\mathcal{D}) = \int P(\bfm{u}|\bfm{f},\bm{w}) P(\bm{w}|\mathcal{D}) \text{d}\bm{w} \approx \int P(\bfm{u}|\bfm{f},\bm{w}) q(\bm{w}|\bm{\theta}^*) \text{d}\bm{w} 
    \label{mc_prediction}
\end{equation}

The resultant predictive distribution can be approximated by Monte Carlo integration of Eq.~(\ref{mc_prediction}) by sampling weights over optimised distributions, $\widetilde{\bm{w}_t}\sim q(\bm{w}|\bm{\theta}^{*})$. At last for a given input force array, $\bm{f}$, probabilistic displacement prediction is obtained as an output. We represent this output distribution by the mean $\mathcal{U}_{\mu}(\bfm{f},\bm{w})$ and the standard deviation $\mathcal{U}_{\sigma}(\bfm{f},\bm{w})$ of the prediction, $P(\bfm{u}|\bfm{f},\mathcal{D})$. This is done by taking mean and standard deviation of $T$ stochastic forwarded passes for the same input as follows:  

\begin{equation}
  \begin{split}
    \mathcal{U}_{\mu}(\bfm{f},\bm{w}) & \approx \frac{1}{T} \sum_{t=1}^{T} P(\bfm{u}|\bfm{f},\widetilde{\bm{w}_t}) \\
    \mathcal{U}^{2}_{\sigma}(\bfm{f},\bm{w}) & \approx \frac{1}{T} \sum_{t=1}^{T} P(\bfm{u}|\bfm{f},\widetilde{\bm{w}_t})^{T}P(\bfm{u}|\bfm{f},\widetilde{\bm{w}_t}) - \mathcal{U}_{\mu}(\bfm{f},\bm{w})^{T}\mathcal{U}_{\mu}(\bfm{f},\bm{w})
   \end{split}
   \label{eq: mean_std_prediction}
\end{equation}

In case of MLE, we do not place distributions over parameters, and they are discrete like in the case of the deterministic network, as in Eq.~(\ref{eq:det_params}). In the penultimate layer, we take ($\bm{d_{\mu}}(\bfm{f},\bm{w})$, $\bm{d_{\rho}}(\bfm{f},\bm{w})$) outputs standing for means and heteroscadetic standard deviations, which are then used to form the final Gaussian distribution output layer. Optimal parameters of the network are computed by minimising the following loss function 

\begin{equation}
      \mathcal{L}_{\text{MLE}}(\mathcal{D,\bm{\theta}_{\text{MLE}}})  \approx  - \sum_{k=1}^{N}\sum_{l=1}^{\mathcal{F}}\left[-\log \left(\sqrt{2\pi}~d^{(l)}_{\sigma}(\bfm{f}^{(k)},\bm{\theta})\right) - \frac{\left(\hat{u}^{(k)}_l- d^{(l)}_{\mu}(\bfm{f}^{(k)},\bm{\theta}_{\text{MLE}})\right)^{2}}{2\left(d^{(l)}_{\sigma}(\bfm{f}^{(k)},\bm{\theta}_{\text{MLE}})\right)^2}\right],
    \label{MLE_loss}
\end{equation}

where
\begin{equation}
         d^{(l)}_{\sigma}(\bfm{f}^{(k)},\bm{\theta}_{\text{MLE}}) = \log(1+\exp({d^{(l)}_{\rho}(\bfm{f}^{(k)},\bm{\theta}_{\text{MLE}}})).
\end{equation}

At last, optimal parameters of MLE U-Net models are computed by minimising the loss functions as

\begin{equation}
      \bm{\theta}_{\text{MLE}}^{*} = \text{arg }\underset{\bm{\theta}_{\text{\tiny MLE}}}{\text{min}}~\mathcal{L}_{\text{MLE}}(\mathcal{D},\bm{\theta}_{\text{\tiny MLE}}) \label{eq:min_VB_MLE}
\end{equation}

%%%%%%%%%%%%%%%%%%%%%%%%%%%%%%%%%%%%%%%%%%%%%%%%
%%%%%%%%%%%%%%%%%%%%%%%%%%%%%%%%%%%%%%%%%%%%%%%%
%%%%%%%%%%%%%%%%%%%%%%%%%%%%%%%%%%%%%%%%%%%%%%%%
\section{Results}\label{sec: Results}

%%%%%%%%%%%%%%%%%%%%%%%%%%%%%%%%%%%%%%%%%%%%%%%%
%%%%%%%%%%%%%%%%%%%%%%%%%%%%%%%%%%%%%%%%%%%%%%%%
\subsection{The numerical experiment procedure}

%%%%%%%%%%%%%%%%%%%%%%%%%%%%%%%%%%%%%%%%%%%%%%%%
\subsubsection{Generation of Training Data from Hyperelastic FEM Simulations}\label{data_generation_FEM}

\begin{figure}[h]
     \centering
     \subfloat[]{\includegraphics[width=0.28\textwidth]{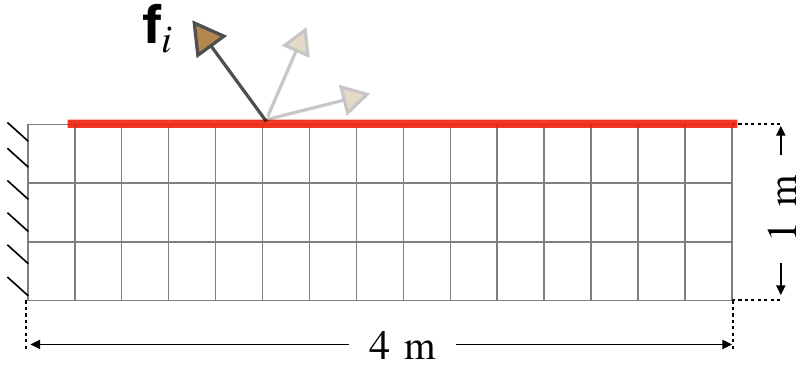}\label{2d_beam_data}}
     \hspace{0.01\textwidth}
     \subfloat[]{\includegraphics[width=0.36\textwidth]{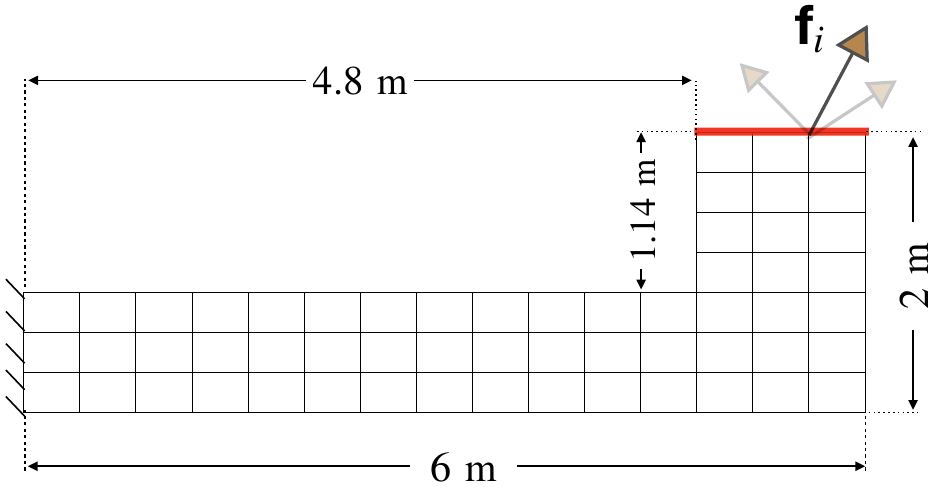}\label{2d_l_data}}
     \hspace{0.01\textwidth}
     \subfloat[]{\includegraphics[width=0.30\textwidth]{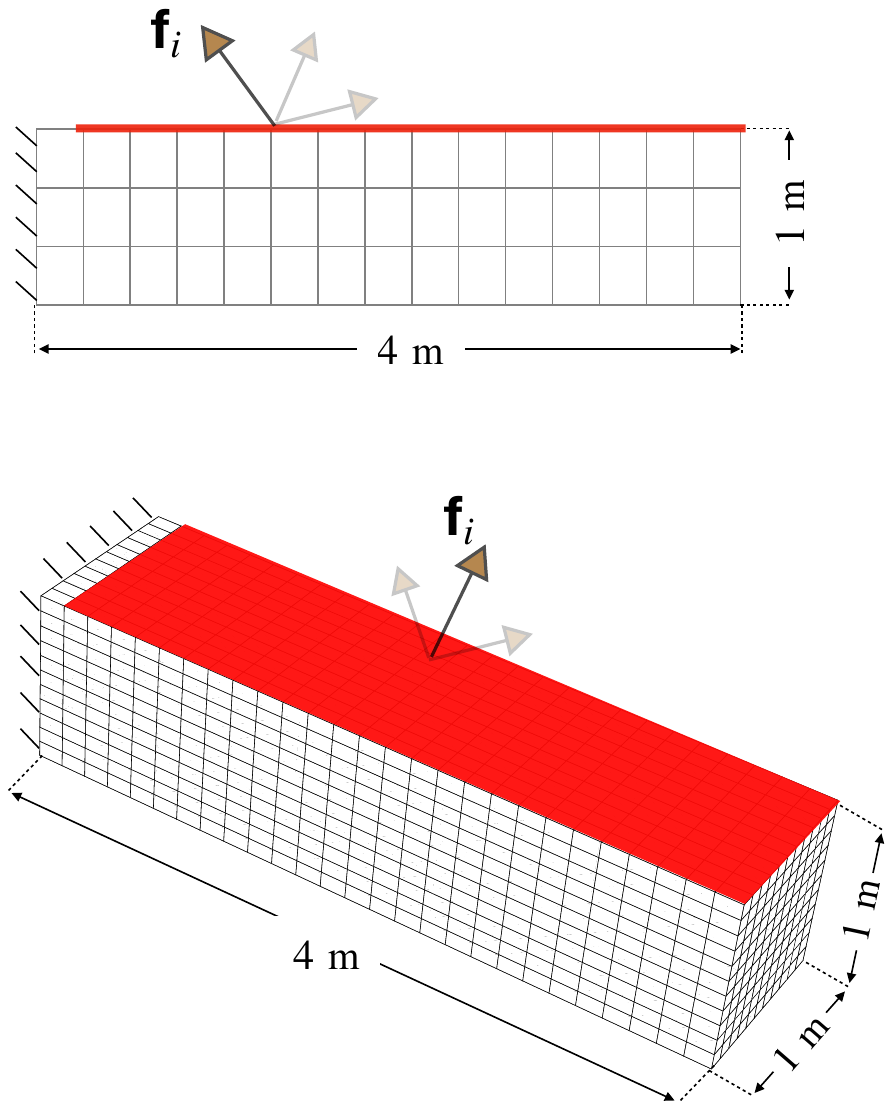}\label{3d_data}}
     \caption{Schematics of three benchmark examples (a) 2D beam, (b) 2D L-shape and (c) 3D beam. The parts of top surfaces marked in red color indicate the nodes at which random nodal forces are applied to generate training datasets.}
     \label{fig: three benchmark examples schematics}
\end{figure}

Two 2D and one 3D benchmark problems are considered in this work, as schematically shown in Figure~\ref{fig: three benchmark examples schematics}. The Neo-Hookean hyperelastic material model is used, with Young's modulus $E=0.5~\text{kPa}$ and the Poisson's ratio $\nu=0.4$. We use the following version of Neo-Hookean strain energy potential 
\begin{equation}
 W(\bfs{F})=\frac{\mu}{2}(I_c-3-2 \ln{J})+\frac{\lambda}{4}( J^2-1-2 \ln{J}),
 \label{Eq:NeoHoohe}
\end{equation}
%OL:SEPEQ1DFHYQ1NeoHooke - confirm with Jakub once 
where the invariants $J$ and $I_{\text{c}}$ are given in terms of deformation gradient $\bfs{F}$ as
\begin{equation}
    J = \text{det}(\bfs{F}), \quad
    I_{\text{c}} = \text{tr}(\bfs{F}^{T}\bfs{F}), \quad
    \text{where} \quad \bfs{F} = \bfs{I} + \nabla \bfs{u},
\end{equation}
while $\mu$ and $\lambda$ are Lame's parameters, which can be expressed in terms of the Young's modulus, $E$, and the Poisson's ratio, $\nu$, as
\begin{equation}
    \lambda = \frac{E\nu}{(1+\nu)(1-2\nu)}, \quad \mu = \frac{E}{2(1+\nu)}.
\end{equation}

As introduced in Section~\ref{sec:fem_based_framework_intro}, for a given discretized problem, the training/testing dataset is constructed as follows. Within nodes occupying a prescribed region of the boundary (in red color in Figure~\ref{fig: three benchmark examples schematics}), a particular family of external force distribution is considered. Each loading case consists of a single excited node, while for the remaining nodes the external forces are $\bfm{0}$. For a given training/testing example, a single node is chosen for which the external force vector is generated randomly, component-wise, from a uniform distribution within a given range of magnitude. The example is then solved with FEM, and the entire vector $\bfm{f}$ of prescribed nodal external forces (including unloaded nodes) and the vector $\bfm{u}$ of computed nodal displacements are saved. The procedure is repeated for all $N+M$ examples, which creates the training/testing dataset $D=\{(\bfm{f}_{(1)},\bfm{u}_{(1)}),...,(\bfm{f}_{(N+M)},\bfm{u}_{(N+M)})\}$.

The finite element simulations have been performed with the AceGen/AceFem framework~\citep{acegen} (standard library displacement-based Neo-Hoohean finite elements are used). The non-linear FE problems are solved with the Newton-Raphson method, and an adaptive load-stepping scheme is used to avoid convergence issues for large load cases. A single quad/hexahedral FE mesh per problem is only considered.

\begin{figure}[h]
     \centering
     \includegraphics[width=\textwidth]{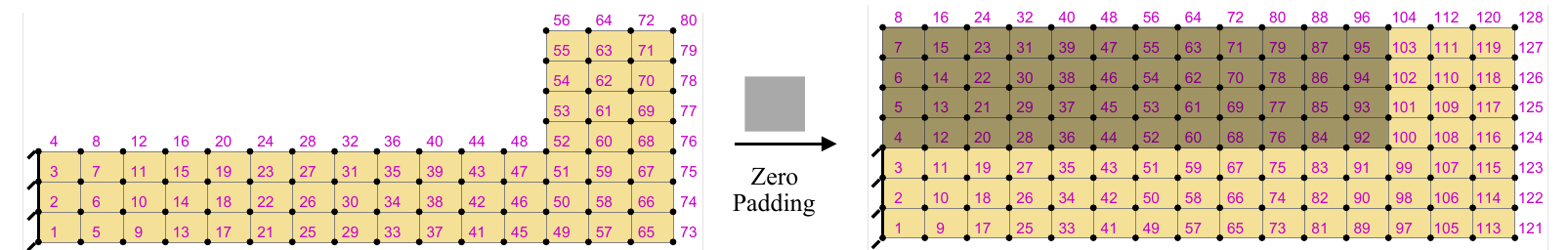}
     \caption{Extra nodes with zero force/displacement are added to convert the data to a structured format. Node numbers are mapped accordingly to follow the assumed order of U-Net architecture.}
     \label{L_padding}
\end{figure}

\emph{Remark:} For 2D/3D beam examples the structured FE mesh is used, which is compatible with the U-Net architecture introduced in Section~\ref{sec:unet_architecture}. In the L-shaped example, the FE mesh is not structured, which makes it impossible to directly transform it to a compatible node numbering, with a possible consequence of accuracy drop, as explained in Section~\ref{dof_order}. To correct this,
a special zero-padding operation is applied to each $\bfm{f}_{(i)}$ and $\bfm{u}_{(i)}$ before using the dataset for training/testing, see Figure~\ref{L_padding}. Note here that unstructured meshes can be handled in several other ways. One way would be to embed a structured grid on the unstructured mesh and map the unstructured nodal values to the structured nodes. Another promising approach would be to use recently developed graph networks \citep{graphcnn}\citep{pfaff2021learning}. These more sophisticated approaches are, however, out of the scope of the present work.  

The datasets are randomly split into training sets, $N$ (95\%), and testing sets, $M$ (5\%). The characteristics of FE meshes and datasets for all three problems are provided in Table~\ref{tab:datasets}.

\begin{table}[h!]
    \centering
    \begin{tabular}{l|c|c|c}
        Problem & N.of FEM DOFs ($\mathcal{F})$ & Force component range [N] & dataset size N+M \\
        \hline
        2D beam & 128 & -2.5 to 2.5 & $5700 + 300$\\
        $\text{2D~beam}^{\#}$ & 128 & -1 to 1 & $3800+200$\\ 
        2D L-shape & 160 ($256^*$) & -1 to 1  & $3800 + 200$\\
        3D beam & 12096 & -2 to 2 & $33688 + 1782$
    \end{tabular}
    \caption{FE datasets. The number of DOFs with the asterisk refers to the zero-padded 2D L-Shaped mesh. $\text{2D~beam}^{\#}$ is an additional dataset used for analysing probabilistic U-Nets in Section~\ref{Sec: Probabilistic results}}
    \label{tab:datasets}
\end{table}

%%%%%%%%%%%%%%%%%%%%%%%%%%%%%%%%%%%%%%%%%%%%%%%%
\subsubsection{Implementation and training of U-Nets}\label{sec:training_UNets}

For the 2D cases, we use 3 level U-Net architectures as in Figure~\ref{fig:CNN-2D}, at each level, we apply two convolutional operators with $3\times3$ filters with c=128 channels in the first level. For Bayesian U-Nets, we replace half of the layers with probabilistic layers (one layer out of two at each level is replaced with a probabilistic layer). For the 3D case, we use a 4 level U-Net architecture, we apply two convolutional operators with $3\times3\times3$ filters with c=128 channels in the first level. Additionally, for both cases, we use batch-normalization on each layer. This technique standardizes the inputs to a layer for each mini-batch \citep{batchnorm}. This has the effect of stabilizing the learning process and dramatically reducing the number of training epochs required to train deep networks.

\textit{Training}: Network is trained by minimising loss function for the given training dataset, minimisation is performed using Adam optimizer, a well-known adaptive stochastic gradient-descent algorithm. We set the learning rate to $1\times10^{-4}$ and set other optimizer parameters as per recommendations~\citep{kingma2017adam}. For the Monte Carlo sampling of loss function ($\mathcal{L}_{\text{VB}}$) in Eq.~(\ref{ELBO_expand}), we use \textit{Flipout} estimator~\citep{wen2018flipout} with its recommended parameter values. Trainings of deterministic and probabilistic versions of U-Net are carried out using Keras~\citep{keras} and Tensorflow-probability~\citep{tfp} libraries respectively. All the implementations are done on Tesla V100-SXM2 GPU, on HPC facilities of the University of Luxembourg \citep{ULHPC} using a batch size of 4 and 600/75 epochs for 2D/3D cases. All the experiments in this work are performed using a single-precision arithmetic ('float32'), which is the usual default choice for all the deep learning libraries. The use of double-precision increased the memory requirements and the training time, without any improvement in the accuracy, and hence is unnecessary. (For the 2D beam example, double precision implementation took 4 times the training time that of the single-precision implementation.)

Since the prediction of Bayesian U-Net is a distribution, we take 300 stochastic forward passes for the same input to get the mean and uncertainty of the prediction (T=300 in Eq.(\ref{eq: mean_std_prediction})).

%%%%%%%%%%%%%%%%%%%%%%%%%%%%%%%%%%%%%%%%%%%%%%%%
\subsubsection{Validation Metrics for the testing phase}\label{validation_metric}
 
For the test set  $\{(\bfm{f}_1,\bfm{u}_1),...,(\bfm{f}_M,\bfm{u}_M)\}$, we use the following mean absolute error norm as the validation metric:

 \begin{equation}
     e_m = \frac{1}{\mathcal{F}}\sum_{i=1}^{\mathcal{F}}{|\mathcal{U}(\bfm{f}_m)^{i} - \bfm{u}_m^{i}|}. 
 \end{equation}
 
$\mathcal{F}$ is the number of dofs of the mesh. For $m^{th}$ test example, $\mathcal{U}(\bfm{f}_m)$ is the deterministic network prediction and $\bfm{u}_m$ is the finite element solution. To have a single validation metric over the entire test set, we compute the average mean norm $\Bar{e}$ and the corrected sample standard deviation $\sigma(e)$ as follows: 

 \begin{equation}
\begin{split}
     \Bar{e} =\frac{1}{M} \sum_{m=1}^{M} e_m, \qquad
     \sigma(e) = \sqrt{\frac{1}{M-1}\sum_{m=1}^{M} \left(e_m - \Bar{e} \right)^2}.
\end{split}
\label{eq:error_metric_det}
\end{equation}

(Note: It is the standard deviation of averaged errors across the test set, not the standard deviation of all errors.) 

In the case of Bayesian U-Nets, the output of the network is a probability distribution, for that reason, we sample over the output distribution by taking multiple forward passes as described in Eq.~(\ref{eq: mean_std_prediction}). Mean over these samples, $\mathcal{U}_{\mu}(\bm{f}_m)$, is taken as the mean prediction of the Bayesian U-Net, while the standard deviation of these samples, $\mathcal{U}_{\sigma}(\bm{f}_m)$, gives us the confidence intervals of predictions. Now the error norm for $m^{th}$ test example is given as 
 \begin{equation}
     e(\mathcal{U}_{\mu}(\bfm{f}_m),\bfm{u}_m) = \frac{1}{\mathcal{F}}\sum_{i=1}^{\mathcal{F}}{|\mathcal{U}_{\mu}(\bfm{f}_m)^{i} - \bfm{u}_m^{i}|} .
 \end{equation}
Again the average error norm and the corrected sample standard deviation for all test examples is computed as 
\begin{equation}
\begin{split}
     \Bar{e} =\frac{1}{M} \sum_{i=1}^{M} e(\mathcal{U}_{\mu}(\bfm{f}_m),\bfm{u_m}), \qquad
     \sigma(e) = \sqrt{\frac{1}{M-1}\sum_{m=1}^{M} \left(e(\mathcal{U}_{\mu}(\bfm{f}_m),\bfm{u}_m) - \Bar{e} \right)^2}.
\end{split}
\end{equation}

%%%%%%%%%%%%%%%%%%%%%%%%%%%%%%%%%%%%%%%%%%%%%%%%
%%%%%%%%%%%%%%%%%%%%%%%%%%%%%%%%%%%%%%%%%%%%%%%%
\subsection{Deterministic U-Nets}\label{sec: deterministic unets}

%%%%%%%%%%%%%%%%%%%%%%%%%%%%%%%%%%%%%%%%%%%%%%%%
\subsubsection{Advantages of the U-Net convolutional architecture}
\label{dof_order}

\textbf{U-Nets vs. fully-connected NNs}

U-Nets leverage the fact that nearby nodes of the FEM mesh show strong local correlation, and provide computationally efficient topology that is able to capture non-linearities. However, if we had to use a fully connected neural network to capture these non-linearities, the number of latent parameters of this network would be significantly larger as compared to that of the U-Net.

To show this effect, we consider the simplest fully connected network, with only input and output layers, without no hidden layers nor activation functions, as a surrogate model for the 3D beam example (as in Figure~\ref{3d_data}). This example has 12096 dof, so the dimension of the input and output layer is 12096 each. Because of the absence of hidden layer/activation functions, this network is only able to capture a linear response of the force-displacement relationship. In order to have the best-linearised approximation, we initialise trainable parameters of the fully connected network with the inverse of the FEM stiffness matrix. Table~\ref{CNN_MLP} shows that the fully connected network (which is an inaccurate assumption) has $1.5$ more parameters than the deterministic U-Net, while the accuracy is greatly reduced. In order to do a better (non-linear) approximation, one would need to use a multi-layer fully connected network, which would require even more parameters, and hence the training time would be significantly higher. Hence, the choice of U-Nets makes complete sense, in particular for complex non-linear problems. 

\begin{table}[h]
\begin{center}
 \begin{tabular}{c | c | c | c } 
 NN type & N. of trainable parameters & $\Bar{e}$ [m] & $\sigma(e)$ [m]\\ [0.5ex] 
 \hline 
 Deterministic U-Net & $~94.1$ E+6 & $0.6$ E-3 & $0.3$ E-3 \\ 
 Fully-connected & $146.3$ E+6 & $7.0$ E-3 & $8.0$ E-3 \\ 
\end{tabular}
\end{center}
\caption{Comparison of U-Net vs feed forward network for 3D Beam example}
\label{CNN_MLP}
\end{table}

\textbf{Effect of DOF ordering}

\begin{figure}[h]
     \centering
     \subfloat[]{\includegraphics[width=0.25\textwidth]{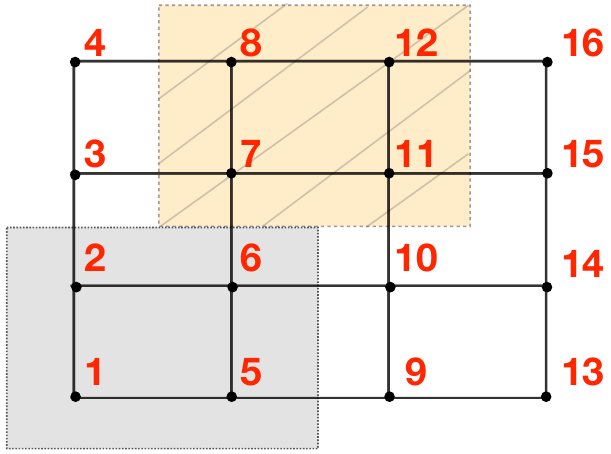}\label{proper_ordering}}
     \hspace{7mm}
     \subfloat[]{\includegraphics[width=0.255\textwidth]{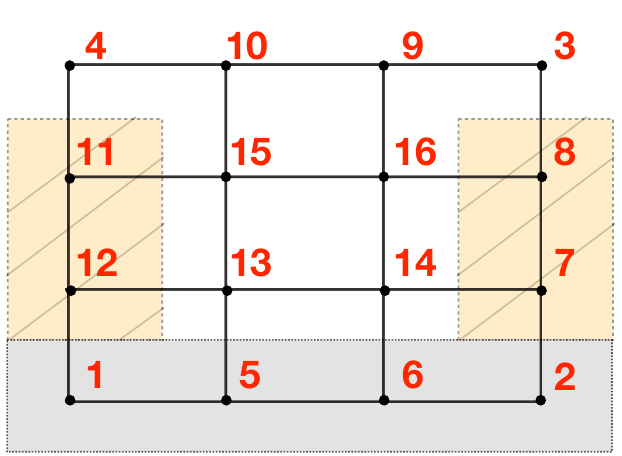}\label{gmsh_order}}
     \hspace{7mm}
     \subfloat[]{\includegraphics[width=0.26\textwidth]{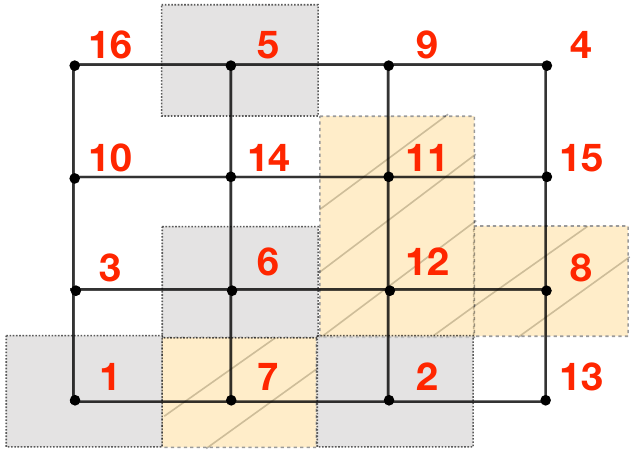}\label{random_ordering}}
     \caption{Different node numbering strategies. (a) Numbering assumed by the TensorFlow (the preferred one; used in this work), (b) Gmsh preprocessor numbering, and (c) random numbering. }
     \label{node_order}
\end{figure}

The topology of input FEM mesh plays a crucial role in the training of U-Nets, and it must be compatible with that of the U-Net architecture topology. However, different FEM pre-processors have different ways of numbering nodes. For instance, Gmsh \citep{Gmsh}, a popular FE mesh generator, first numbers corner nodes, then edge nodes followed by internal nodes, see Figure~\ref{gmsh_order}. This is not compatible with the expected U-Net input, which effects deteriorating the predictive capabilities of the U-Net. A completely random ordering, see Figure~\ref{random_ordering}, performs even worse, see Table~\ref{order_table}. To fully leverage the advantages of U-Nets, care has to be taken to order nodes properly. This is the reason why the zero-padding has been done to the L-shaped case, see the remark in Section~\ref{data_generation_FEM}, and Figure~\ref{L_padding}. \par  

\begin{table}[h!]
\begin{center}
 \begin{tabular}{c | c | c } 
 Ordering strategy & $\Bar{e}$ [mm] & $\sigma(e)~[\text{mm}]$ \\ [0.5ex] 
 \hline
 Preferred ordering (as in Figure~\ref{proper_ordering}) & 0.6 & 0.3\\
 Random ordering (as in Figure~\ref{random_ordering}) & 5.4 & 2.8
\end{tabular}
\end{center}
\caption{Error metrics for the preferred and randomly ordered case for the 3D beam problem.}
\label{order_table}
\end{table}

%%%%%%%%%%%%%%%%%%%%%%%%%%%%%%%%%%%%%%%%%%%%%%%%
\subsubsection{Prediction accuracy}\label{sec: Deterministic}

\begin{figure}[h]
     \centering
     \subfloat[]{\includegraphics[width=0.46\textwidth]{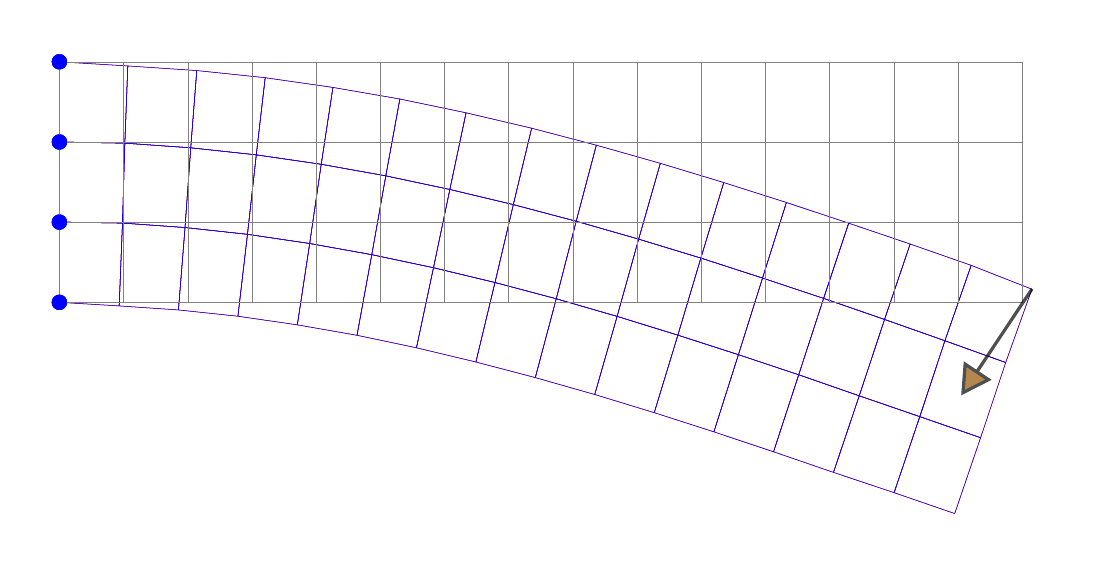}}
     \subfloat[]{\includegraphics[width=0.52\textwidth]{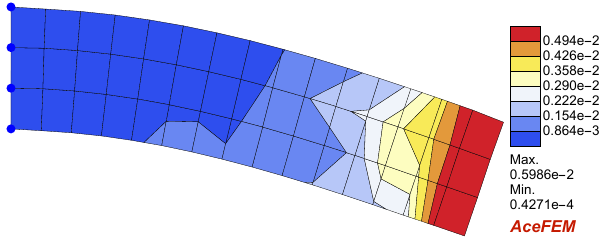}}
     \caption{Deformation of 2D beam computed using the deterministic U-Net, for the point force at the free end. (a) Comparison of deformed meshes, both blue mesh (U-Net solution) and red mesh (reference FEM solution) are overlapping. The magnitude of the tip displacement is 0.95 m. (b) $L_2$ error of nodal displacements between deterministic U-Net and FEM solutions.}
     \label{2d_beam_test}\label{2d_beam_error}
\end{figure}

\begin{figure}[h]
     \centering
     \subfloat[]{\includegraphics[width=0.46\textwidth]{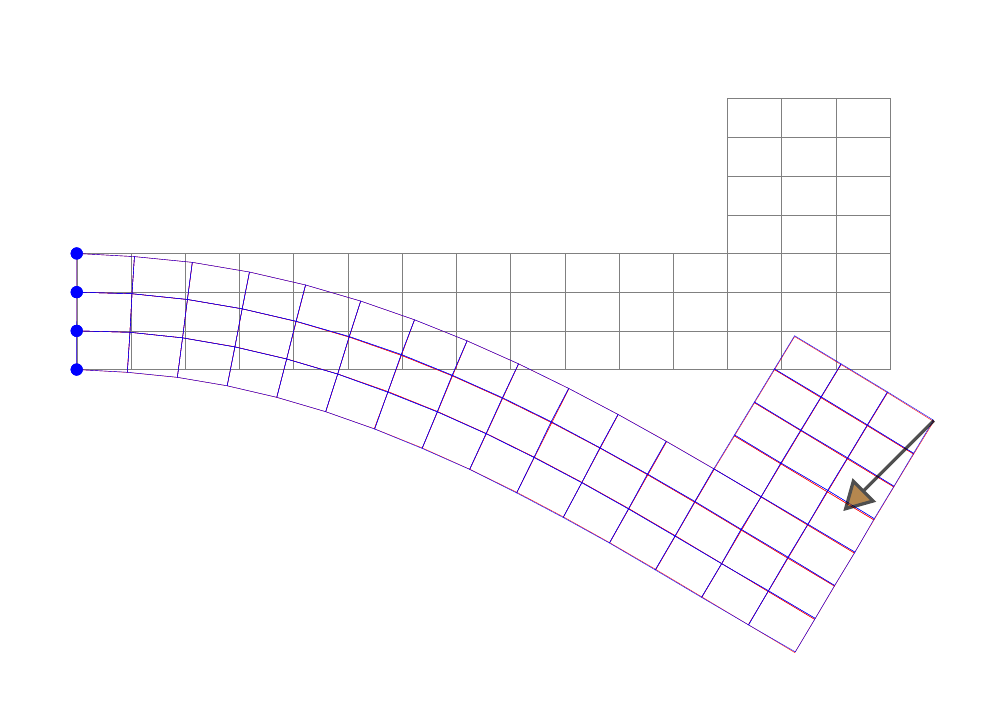}}
     \subfloat[]{\includegraphics[width=0.52\textwidth]{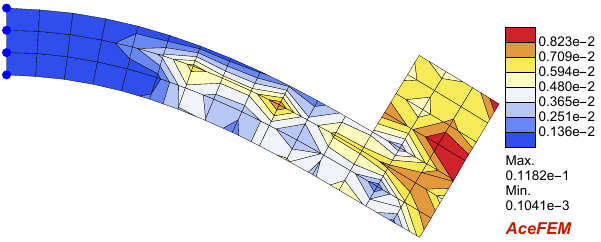}}
     \caption{Deformation of 2D-L shape computed using deterministic U-Net. (a) Initial and deformed meshes predicted using deterministic U-Net(blue) and FEM(red), the magnitude of tip displacement is 2.39 m, and (b) $L_2$ error of nodal displacements between deterministic U-Net and FEM solutions.}
     \label{2d_L_test}\label{2d_L_error}
\end{figure}

\begin{figure}[h]
     \centering
     \subfloat[]{\includegraphics[width=0.45\textwidth]{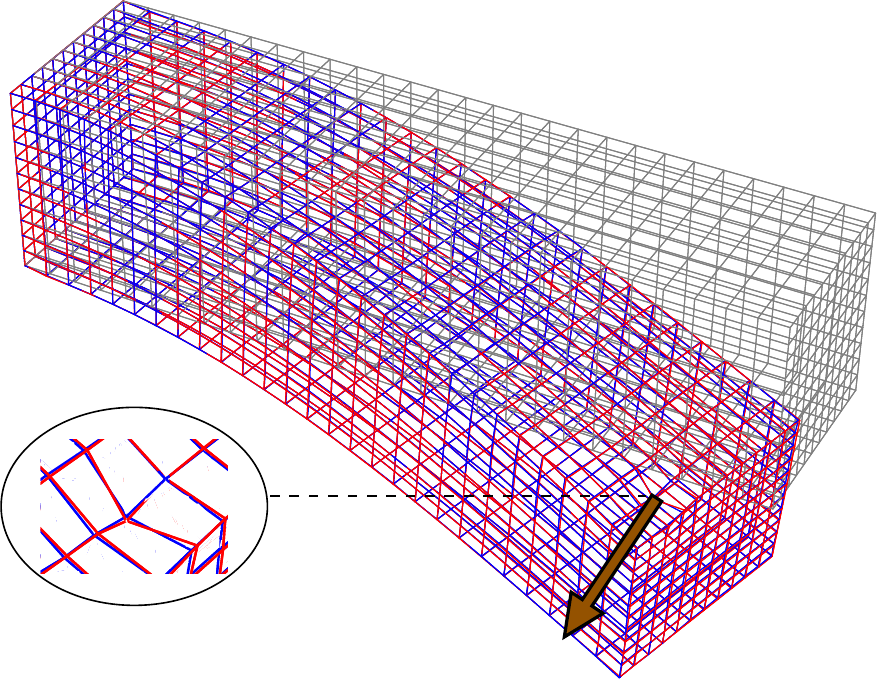}\label{3d1_mesh}}\hspace{0.05\textwidth }
     \subfloat[]{\includegraphics[width=0.4\textwidth]{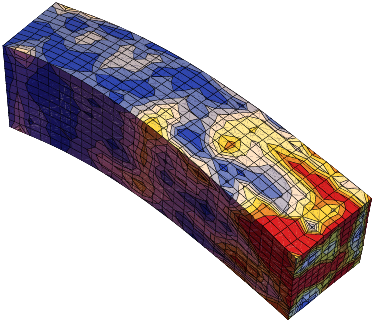}\hspace{0.01\textwidth}\includegraphics[width=0.08\textwidth]{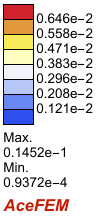}\label{3d1_error}}
     
     \subfloat[]{\includegraphics[width=0.445\textwidth]{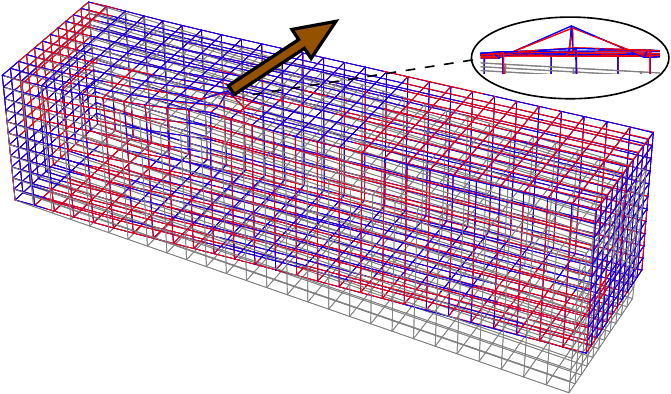}\label{3d2_mesh}}\hspace{0.01\textwidth}
     \subfloat[]{\includegraphics[width=0.435\textwidth]{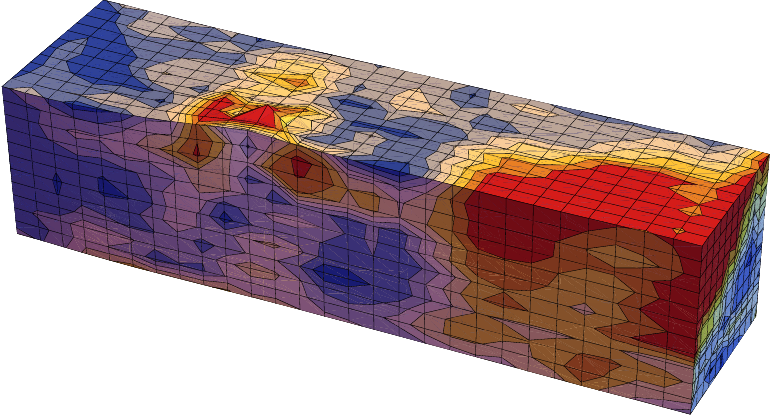}\hspace{0.01\textwidth}\includegraphics[width=0.08\textwidth]{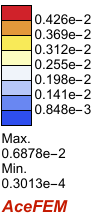}\label{3d2_error}}
     \caption{Deformation of 3D beam computed using deterministic U-Net (blue) for two force cases, for comparison FEM solution (red) is presented. (a)\&(c) deformed meshes for both examples. The magnitude of tip displacement for the first case (force near the free end) is 1.1 m and for the second case (force in the middle region) is 0.26 m. High localized deformations are shown in the insets. (b)\&(d) $L_2$ error of nodal displacements between deterministic U-Net and FEM solutions.} 
     \label{fig: 3d_test}
\end{figure}

\begin{figure}[h]
     \centering
     \includegraphics[scale=0.5]{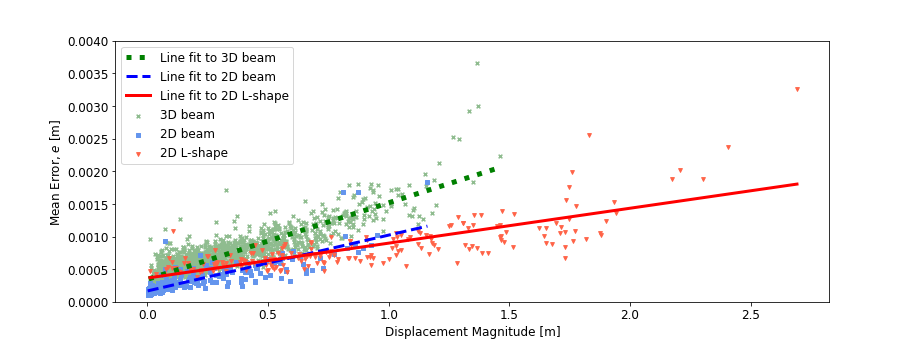}
     \caption{Mean errors ($e$) for all test examples of three benchmark cases. The regression lines $y \propto 0.0008 \times x$ (2D-beam),  $y \propto 0.0005 \times x$ (2D-L shape),  $y \propto 0.001 \times x$ (3D beam) show low sensitivity of the deterministic U-Nets to displacement magnitudes.}
     \label{fig: det_sensitivity}
\end{figure}

\begin{figure}[h]
     \centering
     \subfloat[]{\includegraphics[width=0.46\textwidth]{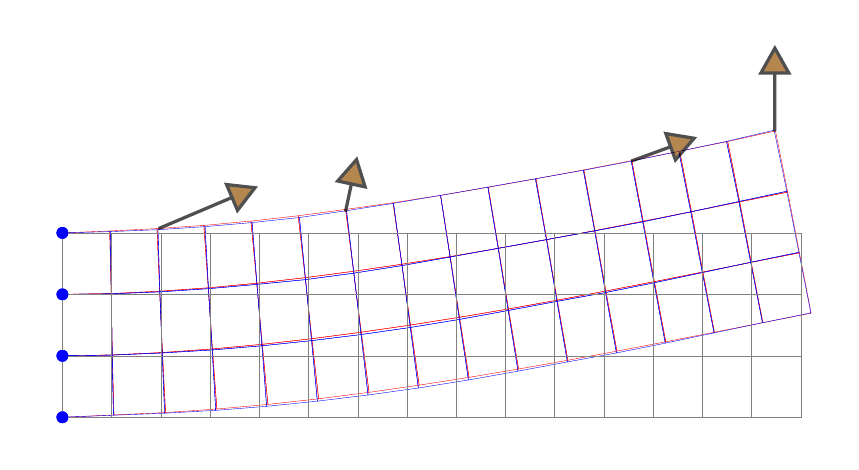}}
     \subfloat[]{\includegraphics[width=0.52\textwidth]{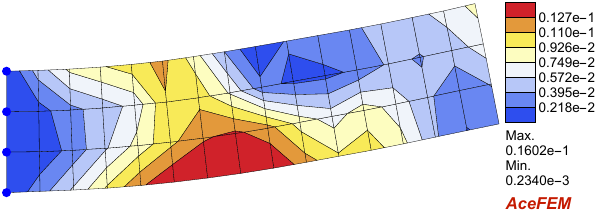}\label{fig: error_multiple}}
     \caption{Deformation of the 2D beam subjected to multiple point loads. (a) Comparison of deformed meshes predicted with deterministic U-Net and FEM, the magnitude of tip displacement is 0.55 m. (b)  $L_2$ error of nodal displacements between deterministic U-Net and FEM solution.}
     \label{fig: multiple_force}
\end{figure}

Deterministic U-Nets are trained on FEM datasets generated as described in Section~\ref{data_generation_FEM}. Below, we analyse in a more detail some selected test examples for each case, and compare their FEM and U-Net solutions. For all the examples, we show the overlap of deformed meshes obtained using FEM and U-Net models. In Figures~{\ref{2d_beam_test}-\ref{fig: multiple_force} and Figure~\ref{fig: training_away}}, gray, blue and red meshes represent undeformed configuration, U-Net solution and FEM solution, respectively. In addition to that, we also present the interpolated node-wise $L_2$ norm of the prediction error (the error of the nodal displacement between FEM and U-Net. 

In Figure~\ref{2d_beam_test}, we show a test example of the 2D-beam case. A point force is applied at the corner node of the beam and the deformation of mesh is predicted using the deterministic U-Net. As we can see, the deformed mesh predicted with U-Net is overlapping with the reference FEM solution. As explained above, we also plot the nodal error field on the deformed mesh, one can observe that the error is relatively higher in the high displacement region, i.e, near the free end. The relative error for the tip with respect to its displacement magnitude is only 0.6\%.

Figure~\ref{2d_L_test} shows an example of the 2D L-shape case. Again the deterministic U-Net solution is overlapping with the reference FEM solution. $L_2$ error contour shows that a high error trend is observed near the free end again, the relative error at the top corner node with respect to displacement magnitude is 0.4\% only.

We further take a look at two 3D-beam test examples, one with the force applied near the free end and another with the force applied in the middle region of the 3D beam. For both cases, the deterministic U-Net solutions are overlapping with the FEM solutions. Insets in Figure~\ref{fig: 3d_test} show that the U-Net is capable of predicting high local non-linear deformations. For the first example in Figure~\ref{3d1_mesh}-\ref{3d1_error}, the error field shows high error region near the point of application of the force. The relative error for the tip for this case is only 0.6\%. Whereas, Figure~\ref{3d2_mesh}-\ref{3d2_error} shows an example with the force applied relatively near to the fixed end. In this case, a high error field is observed at the point of application of force as well as near the free end. The relative $L_2$ error of the tip for this example is 1.6\%. From this, we can say that errors are usually higher near the nodes with higher displacement magnitudes. 

Till now we looked at the prediction accuracy for individual examples, now we would like to judge the performance of deterministic U-Net over the entire test sets (5\% of the generated data is designated for testing purposes). Table~\ref{tab: det_metrics} provides such comparison in a form of averaged error over entire test sets. We can see that, on average, the error is at a reasonably low level. To extend this analysis, in Figure~\ref{fig: det_sensitivity} we plot the mean error ($e$) of each test example of the three benchmark problems. We sort these errors as per the increasing displacement magnitude at the point of application of force. To get a relation between displacement and mean error ($e$), we do a least square linear fit for all three cases. From Figure~\ref{fig: det_sensitivity}, all the three examples show generally low sensitivity to the increase of displacement magnitude.

\begin{table}[h]
\begin{center}
 \begin{tabular}{l | c | c | c   } 
 Example & $M$ & $\Bar{e}$ [m] & $\sigma(e)~[\text{m}]$ \\ [0.5ex] 
 \hline
 2D Beam & 300 & 0.3 E-3 & 0.2 E-3 \\ 
 2D L-Shaped & 200 & 0.8 E-3 & 0.4 E-3\\ 
 3D Beam & 1782 & 0.6 E-3 & 0.3 E-3 
\end{tabular}
\end{center}
\caption{Error metrics for 2D and 3D test sets for predictions using deterministic U-Net. $M$ stands for the number of test examples, and $\Bar{e}$ and $\sigma(e)$ are error metrics defined in Section~\ref{validation_metric}.}
\label{tab: det_metrics}
\end{table}

\textbf{Effect of changing the distribution of applied forces}

Deterministic U-Net has been trained by using single point load examples only, but we would like to check how it performs when multiple point load input is given for the prediction. Figure~\ref{fig: multiple_force} shows one such example where random multiple forces are applied on the top edge, U-Net is able to closely follow the reference FEM solution. Figure~\ref{fig: error_multiple} shows the $L_2$ norm of the error across the beam, it shows a different trend for this example. Though the deformation is higher in the free end region and at the point of application of forces, a higher error is observed at a different location also. Solution accuracy of multiple point load cases can be improved by incorporating multiple point loads in the training phase. Also, the relative error for the tip with respect to its displacement magnitude for this example is 0.6\%.

In most engineering applications, we are interested in the cases where force is applied in a given prescribed region of interest (e.g., the Neumann boundary). Here, we would like to check how U-Net performs when this assumption is violated, i.e, we apply forces on the nodes which were not involved in the training procedure. To do so, we use the same 2D beam case with the training set generated by applying point forces on the top edge (indicated by the red line in Figure~\ref{fig: three benchmark examples schematics} in Section~\ref{data_generation_FEM}). What we change is the prediction phase, during which we apply forces on nodes located on the vertical free edge of the beam (see schematics in Figure~\ref{fig: training_away}). In the example, we apply a vertical force of 1.5 N on each of the 4 nodes of the free edge of the beam. Figure~\ref{fig:away_prediction} shows that mesh (blue) predicted with U-Net deviates more and more from the true FEM solution, as we move away from the training line. The U-Net solution is much worse when the force is applied on the $4^{\text{th}}$ node as compared to the $2^{\text{nd}}$, i.e. when the point of force application is farthest from the training line. In Figure~\ref{fig:away_errors}, we plot the mean and maximum errors of all 4 examples, and we can observe a significant accuracy drop reaching two orders of magnitude when predicting outside the training region. Also, we can see that the errors are increasing when moving away from the training dataset. This proves that the U-Nets extrapolate predictions poorly when moving away from the training range in spatial directions.

\textbf{Training convergence}

\begin{figure}[h]
     \centering
     \subfloat[]{\includegraphics[width=0.5\textwidth]{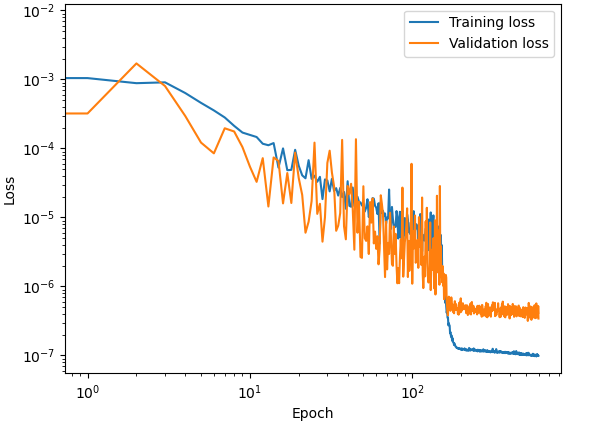}}
     \subfloat[]{\includegraphics[width=0.5\textwidth]{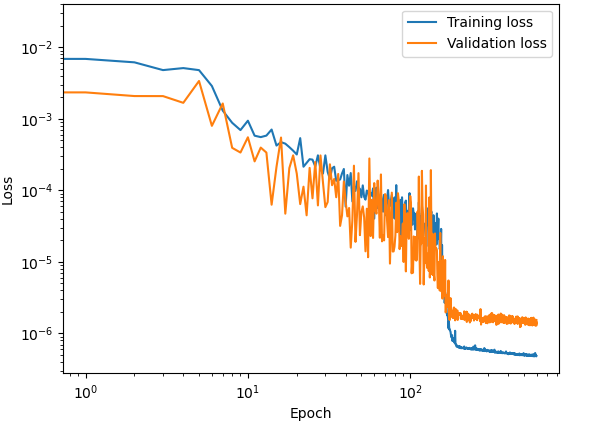}\label{fig: loss l}}
     \caption{Mean squared error log-loss plots for (a) 2D beam and (b) 2D L-shape case.}
     \label{fig: loss}
\end{figure}

\begin{figure}[h]
     \centering
     
     \subfloat[]{\raisebox{0.00\textwidth}{\includegraphics[width=0.4\textwidth]{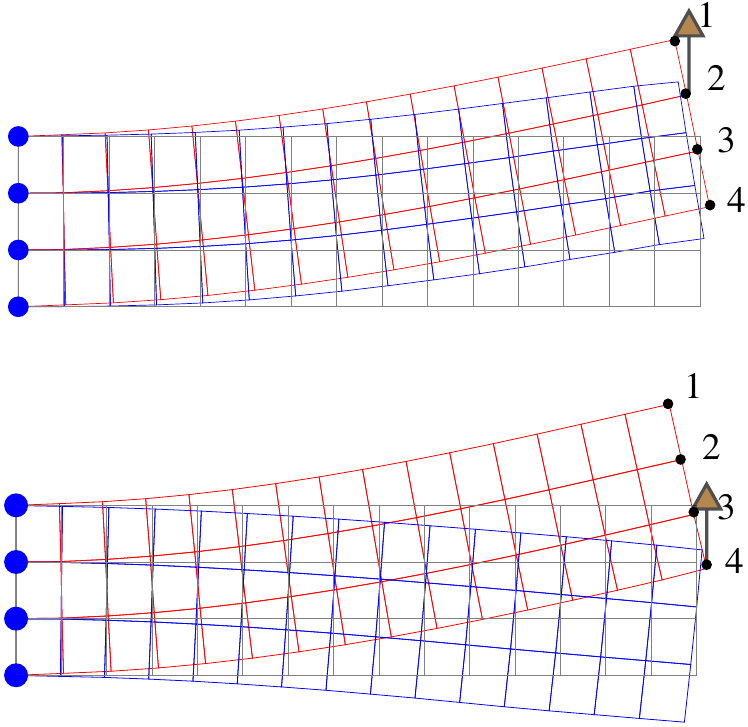}}\label{fig:away_prediction}}
     \subfloat[]{\includegraphics[width=0.6\textwidth]{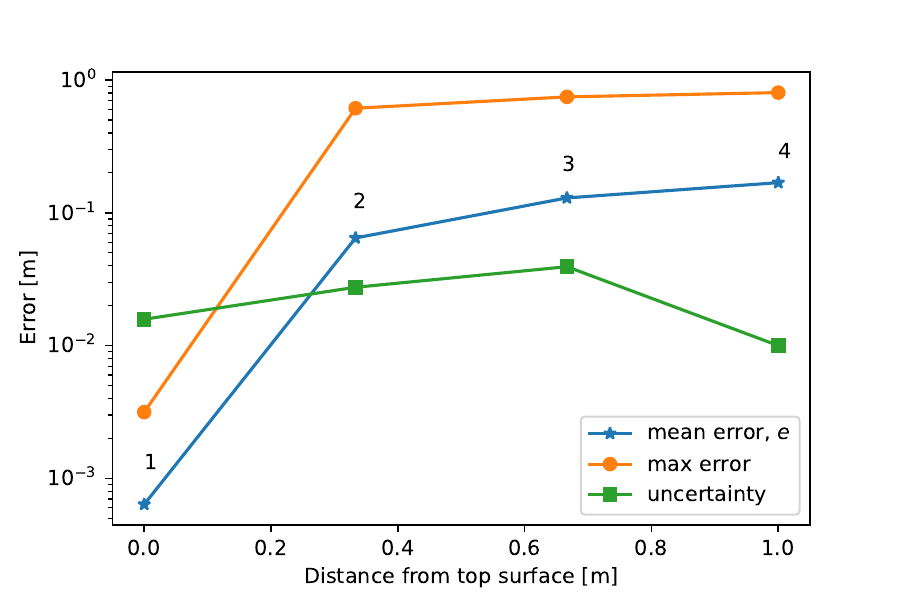}\label{fig:away_errors}}
     \caption{Application of point loads on the nodes away from the training region. (a) Deformed meshes obtained by FEM (red) and with U-Net (blue) when point load is applied on $2^{\text{nd}}$ or $4^{\text{th}}$ node. (b) Mean error ($e$) and maximum error for each of the 4 point loads cases. Green line shows the uncertainty prediction of Bayesian U-Net, for the node on which the force is applied.}
     \label{fig: training_away}
\end{figure}

The choice of the amount of training data and the appropriate neural network architecture are two important criteria in the case of neural network surrogate modeling. This is crucial to ensure that neither underfitting nor overfitting is observed. For all the cases in this work, training convergence is ensured by observing loss plots of training and validation errors, i.e., the training error doesn't decrease, and validation error doesn't go higher with the number of epochs. For the reference, the loss plots for 2D cases are shown in Figure~\ref{fig: loss}.

\subsection{Probabilistic U-Nets}\label{Sec: Probabilistic results}

The goal of our probabilistic U-Net framework is to get reliable predictions and uncertainty associated with those predictions. Further in this section, we will check this for the case of data and model uncertainties for selected examples analogous to the deterministic case.

\subsubsection{Prediction accuracy}\label{sec: Pred_Bayes_UNET}

We train the probabilistic U-Net framework on the same datasets as used in deterministic cases. Because the output of the network is a distribution, we make 300 stochastic forward passes to get the mean and uncertainty predictions for a given input. Mean prediction of Bayesian U-Net is treated as the solution of the network, whereas uncertainty predictions give information of credible intervals of predictions. Table~\ref{tab:bayesian_table} gives the error metrics for the Bayesian U-Net predictions over the entire test sets, for comparison we have shown the errors of deterministic counterparts as well. 

\begin{table}[h]
\begin{center}
 \begin{tabular}{l | c | c | c } 
 Example & $M$ & $\Bar{e}~[\text{m}]$ & $\sigma(e)~[\text{m}]$ \\[0.5ex] 
 \hline
2D Beam (VB)& 300 & 1.3 E-3 & 1.3 E-3  \\
2D Beam (D)& 300 & 0.3 E-3 & 0.2 E-3  \\
\hline
2D L-Shaped (VB) & 200 & 5.3 E-3 & 3.7 E-3  \\
2D L-Shaped (D)& 200 & 0.8 E-3 & 0.4 E-3 

\end{tabular}
\end{center}
\caption{Error metrics for 2D test sets using Bayesian U-Nets. D = Deterministic, VB = Variational Bayes.}
\label{tab:bayesian_table}
\end{table}

\begin{figure}[h]
     \centering
     \includegraphics[scale=0.5]{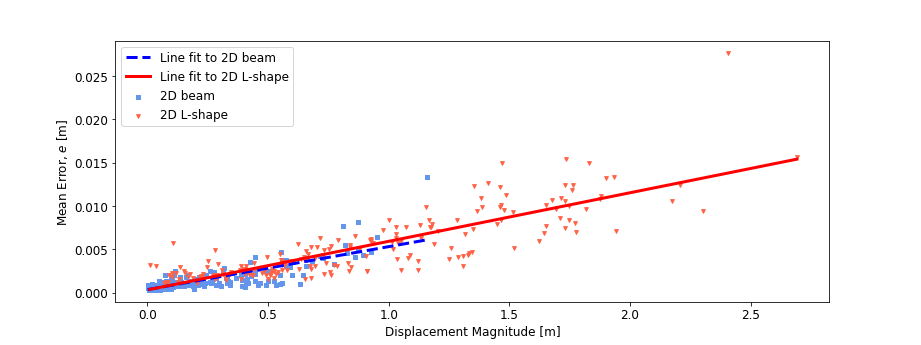}
     \caption{Mean errors ($e$) for all test examples of 2D cases, for predictions using Bayesian U-Net. The regression lines $y \propto 0.005 \times x$ (2D beam), $y \propto 0.006 \times x$ (2D-L shape) show low sensitivity of Bayesian U-Net errors to displacement magnitudes.}
     \label{average_errors_bayesian}
\end{figure}

Similar to the deterministic case, we do the analysis of the error metric ($e$) for all the test examples predicted using Bayesian U-Net this time. Figure~\ref{average_errors_bayesian} shows errors sorted as per the increasing displacement magnitudes of the point of application of forces. We perform a least-squares line fit to the error data. Slopes for 2D-beam and 2D L-shape cases are small, proving a little sensitivity of errors to the displacement magnitudes.  

Hereafter we focus on particular examples to get more insights on Bayesian U-Net predictions. Similar to the deterministic cases, we take node-wise $L_2$ norm of the error (Error of FEM and mean prediction of Bayesian U-Net) and also that of the uncertainty prediction from Bayesian U-Net. Both error and uncertainty values are interpolated within the element to get respective fields, which are plotted on the deformed mesh obtained using Bayesian U-Net.  

\begin{figure}[h]
     \centering
     \subfloat[]{\includegraphics[width=0.48 \textwidth]{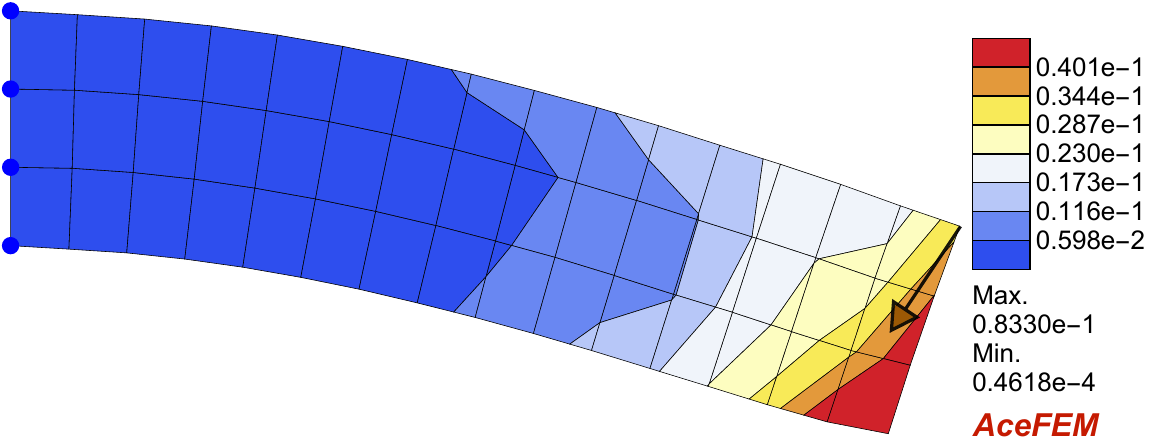}\label{2d_beam_bayes_error}}
     \subfloat[]{\includegraphics[width=0.48 \textwidth]{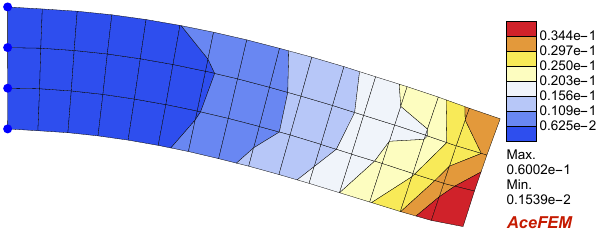}\label{2d_beam_bayes_uncertainty}}
     \caption{Deformation of 2D Beam predicted by the Bayesian U-Net, for the same example as in Figure~\ref{2d_beam_test}. (a) The error between Bayesian U-Net and FEM solution plotted on the deformed mesh. (b) Uncertainty of prediction obtained using Bayesian U-Net plotted on the deformed mesh.}
     \label{fig: 2d_beam_error_uncertainty}
\end{figure}

We consider the same 2D-beam test case as in Figure~\ref{2d_beam_test}, (as in deterministic case). This time we make the prediction using Bayesian U-Net. Figure~\ref{fig: 2d_beam_error_uncertainty} shows the comparison of error and uncertainty associated with the prediction (we plot single standard deviation values associated with the prediction of respective dof). One can see that both are strongly co-related spatially.

\begin{figure}[h]
     \centering
     \subfloat[]{\includegraphics[width=0.48 \textwidth]{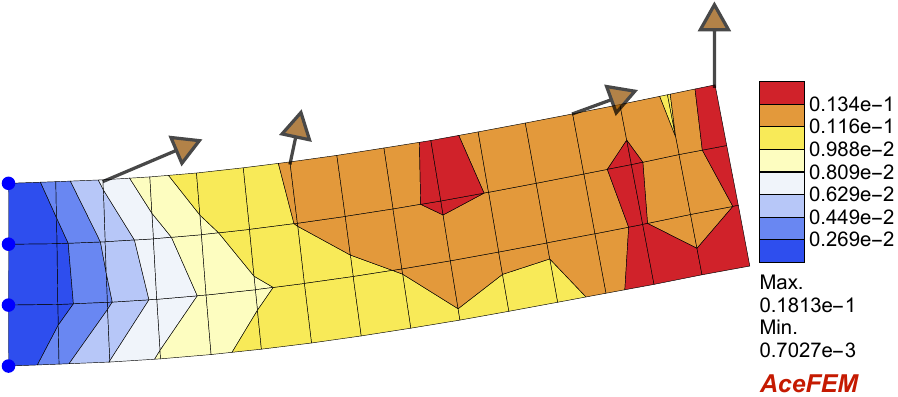}\label{2d_multiple_bayes_error}}
     \subfloat[]{\includegraphics[width=0.48 \textwidth]{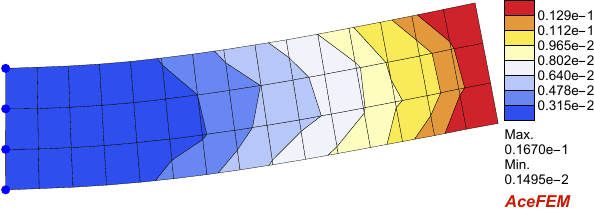}\label{2d_multiple_bayes_uncertainty}}
     \caption{Deformation of 2D Beam for multiple point forces using Bayesian U-Net, for the same example as in Figure~\ref{fig: multiple_force}. (a)  The error between Bayesian U-Net and FEM solution plotted on the deformed mesh. (b) Uncertainty of prediction obtained using Bayesian U-Net plotted on the deformed mesh.}
     \label{fig: 2d_multiplef_error_uncertainty}
\end{figure}

A similar kind of analysis is done for the multiple point load case Figure, see~\ref{fig: multiple_force}, in the deterministic section. Figure~\ref{fig: 2d_multiplef_error_uncertainty} compares the error and uncertainty fields obtained using the Bayesian U-Net, we can see that they are correlated and closely follow each other as well.

\begin{figure}[h]
     \centering
     \subfloat[]{\stackinset{r}{.122\textwidth}{t}{-.0475\textwidth}{\includegraphics[width=0.02\textwidth]{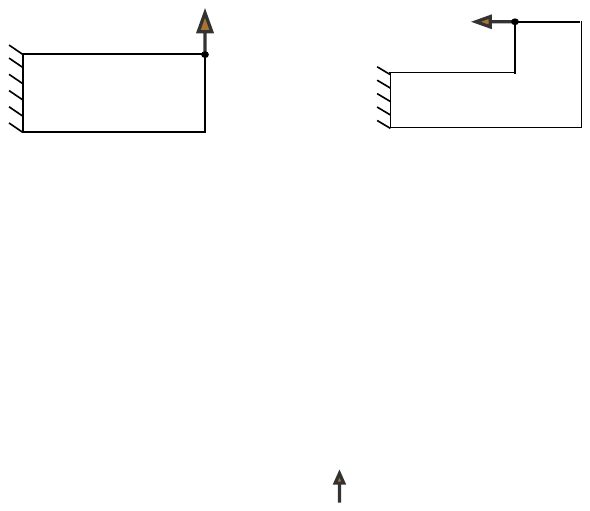}}{\includegraphics[width=0.48 \textwidth]{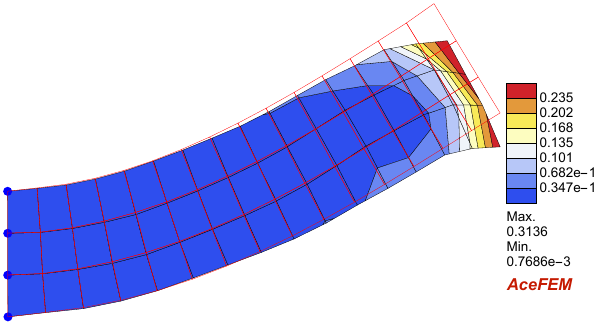}\label{2d_5Nf_bayes_error}}}
     \subfloat[]{\includegraphics[width=0.48 \textwidth]{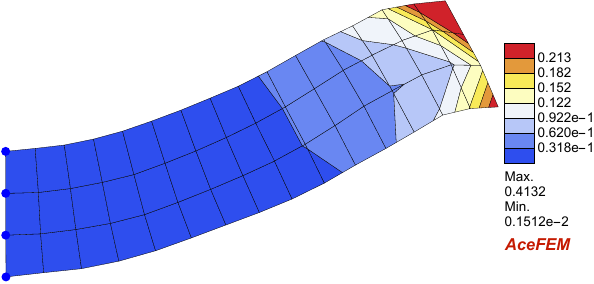}\label{2d_5Nf_bayes_uncertainty}}
     \caption{Deformation of 2D Beam using Bayesian U-Net for an input force outside the training range. (a)  Error between Bayesian U-Net and FEM solution plotted on the deformed mesh. (b) Uncertainty of prediction obtained using Bayesian U-Net plotted on the deformed mesh.}
     \label{fig: 2d_5Nf_error_uncertainty}
\end{figure}

Force outside training range: Let us consider a test example in which a force of 5~N is applied on the corner node, which is far away from the training range (which is -2.5 to 2.5 N). Again we compare error and uncertainty associated with the Bayesian U-Net prediction. For reference, the FEM solution is presented (red mesh) with the error contour plot. Both error and uncertainty are plotted on the deformed mesh predicted with the Bayesian U-Net. In Figure~\ref{fig: 2d_5Nf_error_uncertainty}, one can see that both are strongly correlated, rather both values are close to each other across the spatial dimensions of the beam. Thus, the uncertainty predictions can give us an idea about the error of U-Net predictions, irrespective of whether an input is within or outside the training region. 

For each of the above examples shown in Figure 14-16, we can see that the U-Net solution is deviating from the true FEM solution, which is given by the error contour, i.e., the U-Net model is not able to fit the data exactly. And uncertainty prediction obtained using the Bayesian U-Net is able to capture this fitting error.

The example in Figure~\ref{fig: 2d_5Nf_error_uncertainty} can be considered as a case of extrapolation in the sense of the magnitude of force being outside the training range. One can also think of extrapolation in the sense of applying force on the nodes which were not included in the training, i.e., extrapolation in the spatial dimensions of the geometry. To analyse such cases we consider the same example as shown in Figure~\ref{fig: training_away} in the deterministic Section~\ref{sec: Deterministic}. In Figure~\ref{fig:away_errors} we have shown the uncertainty of Bayesian U-Net prediction (one standard deviation), for the node on which point load is applied. As one can infer, Bayesian U-Net is not giving reliable uncertainty estimates when we move away from the training region in spatial directions. Intuitively the predicted uncertainty should be more for the case when force is applied on the farthest node from the training line, but on contrary, we observed a low prediction uncertainty for this point. One possible explanation of this limitation is, gradients w.r.t the spatial dimensions are not available neither in the data nor in the U-Net models. Hence there is no natural way of extrapolating information of solutions or uncertainties.

\begin{figure}[h]
     \centering
     \subfloat[]{\stackinset{r}{.29\textwidth}{t}{.05\textwidth}{\includegraphics[width=0.12\textwidth]{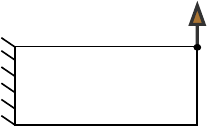}}{\includegraphics[width=0.49 \textwidth]{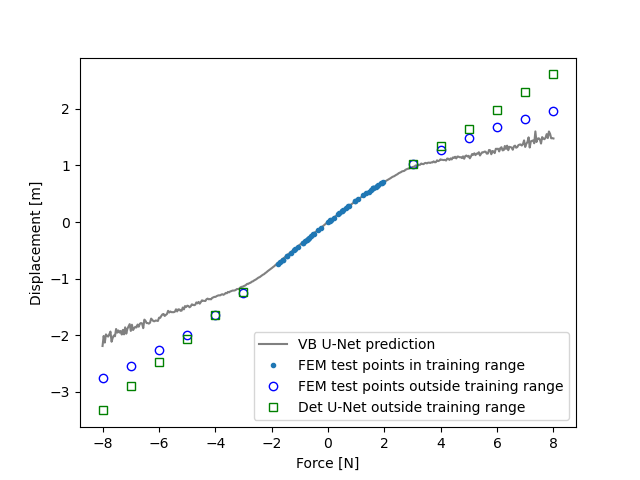}\label{mean_bar2.5}}}
     \subfloat[]{\includegraphics[width=0.49 \textwidth]{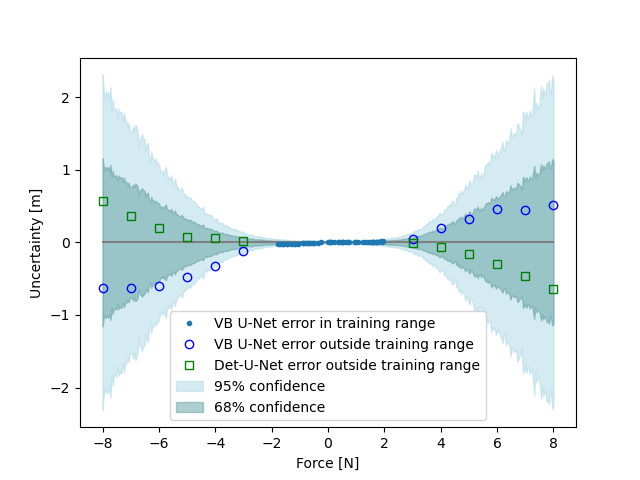}\label{std_bar2.5}}

     \caption{Outputs of Bayesian U-Net when a range of forces is applied on the corner node of the 2D Beam (see inset). (a) The magnitude of Y-displacement of the corner node predicted with Variational Bayesian U-Net. (b) The uncertainty associated with the predictions of displacement solutions.}
     \label{fig: fa2half_deform}
\end{figure}

\begin{figure}[h]
     \centering
     {\stackinset{r}{.185\textwidth}{t}{0.05\textwidth}{\includegraphics[width=0.12\textwidth]{Images/inset_bar.pdf}}
     {\includegraphics[width=0.5 \textwidth]{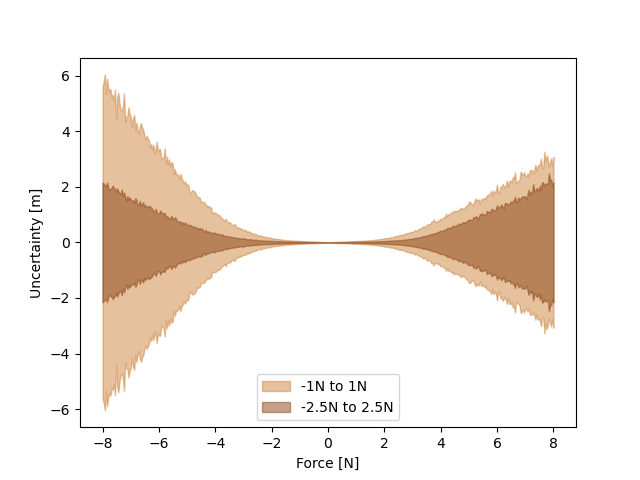}}}
     \caption{Uncertainty intervals for the Y-displacement of the corner node of the 2D beam (see inset), predicted using the Bayesian U-Nets trained on different training sets. Uncertainty reduces with the increase of force range in training data.}
     \label{overlap_intervals}
\end{figure}

\begin{figure}[h]
     \centering
     \subfloat[]{\stackinset{r}{.32\textwidth}{t}{.055\textwidth}{\includegraphics[width=0.10\textwidth]{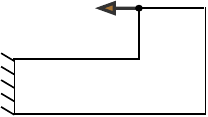}}{\includegraphics[width=0.49 \textwidth]{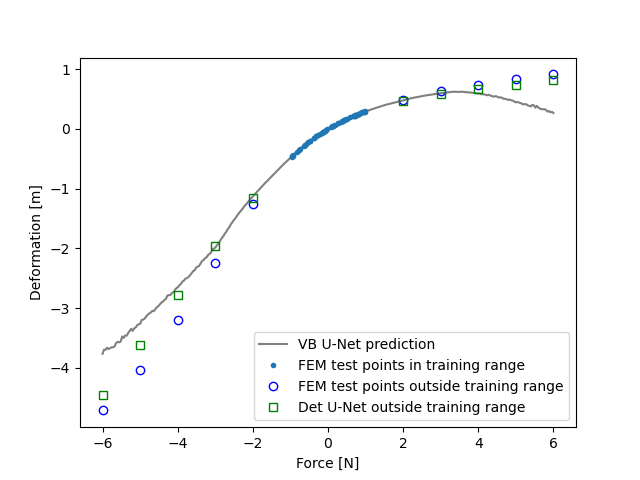}\label{mean_L2.5}}}
     \subfloat[]{\includegraphics[width=0.49 \textwidth]{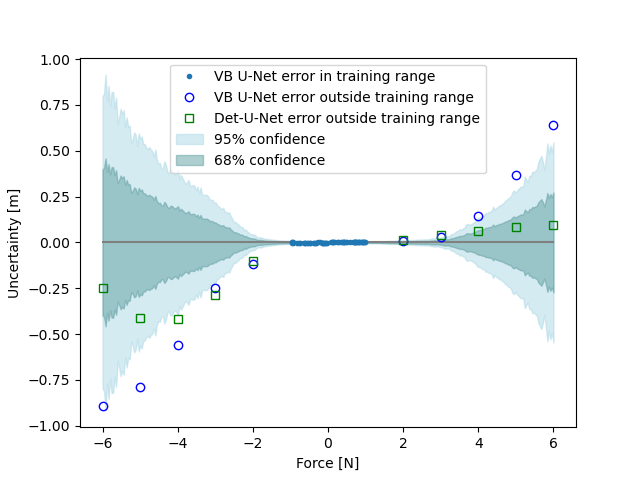}\label{std_L2.5}}

     \caption{Outputs of Bayesian U-Net when a range of forces is applied on the inner corner node of the 2D L-shape (see inset). (a) The magnitude of X-displacement of the inner corner node predicted with Variational Bayesian U-Net. (b) The uncertainty associated with the predictions of displacement solutions.}
     \label{fig: L_deform}
\end{figure}

Hereafter we focus on displacement prediction of a single dof with Bayesian U-Net. We do this to see how the associated uncertainty varies with the value of input force, depending on whether the input is within or outside the training range. In Figure~\ref{fig: fa2half_deform}-\ref{fig: MLEvsBayes}, we keep a constant direction of the input force, but gradually increase the magnitude and study the displacement prediction using Bayesian U-Net. The output of the Bayesian U-net is the displacement solution and the uncertainty associated with it, in Figure~\ref{fig: fa2half_deform}-\ref{fig: L_deform} we provide separate plots for both of these outputs. Whereas in Figure~\ref{fig: MLEvsBayes} we only study the prediction uncertainties for noisy data cases.  

2D Beam: We apply multiple vertical forces varying from -8~N to 8~N on the corner node of the beam and predict its displacements using the Bayesian U-Net. Figure~\ref{mean_bar2.5} gives the prediction of displacement magnitude, as one can see prediction matches with test FEM solution within the training region. Outside the training region, Bayesian U-Net prediction deviates from the FEM solution. For reference, we provide deterministic U-Net solutions as well, even they deviate from FEM solutions outside the training range.  Whereas Figure~\ref{std_bar2.5} gives confidence intervals associated with these predictions. One can see that network has very little uncertainty i.e. it is confident in the region of training data (-2.5 to 2.5 N) but as one moves away, the uncertainty of the prediction increases. We can also see that 95\% confidence is able to capture the error of Bayesian U-Net predictions outside the training region, for reference, errors of deterministic U-Nets are presented as well. 

In this paragraph, we compare uncertainty intervals for Bayesian U-Nets trained on two different datasets. In addition to the existing 2D beam dataset (force range: -2.5 to 2.5 N), we consider another training dataset with a lower force range this time (force range: 1 to 1 N). Figure~\ref{overlap_intervals} shows the comparison of uncertainty intervals for these two cases, as the range of input force in the training set is decreasing, Bayesian U-Net tends to get more uncertain about its predictions in higher force ranges, which follows the common intuition.   

2D L-shape: This training dataset was created by applying point forces in the range of -1 N to 1 N as shown in Table~\ref{tab:datasets}.  In order to see how prediction uncertainty varies with the input forces, we apply multiple forces in a horizontal direction varying from -6 N to 6 N on the inner corner of the L-shape and predict its displacements using Bayesian U-Net. Figure~\ref{mean_L2.5} shows how displacement magnitude changes with applied force values. As we start to move away from the training region, the Bayesian U-Net solution deviates from the FEM solution. For reference, we have plotted the deterministic U-Net solutions as well. Figure~\ref{std_L2.5} gives the uncertainty associated with the prediction. Again the network is very confident in the training data region. But as the force value goes outside the training range, uncertainty tends to increase, for the reference, errors of deterministic U-Nets are presented as well.    

\subsubsection{Noisy Data Case}

In all the cases above, the U-Net models have been trained with numerical FEM datasets which can be regarded as noiseless. However, in many practical applications, especially when working with experimental data, the data noises exist and can originate from various sources, such as measurement errors, errors associated with tools, human errors, etc. In this section, we would like to demonstrate that our framework is capable of capturing these data noises. To show that, we add random noises to our existing FEM datasets, and check how MLE- and Variational Bayes U-Nets perform in capturing these noises in terms of the predicted uncertainties.

\begin{figure}[t!]
     \centering
     \subfloat[]{\stackinset{r}{.31\textwidth}{t}{.05\textwidth}{\includegraphics[width=0.10\textwidth]{Images/inset_bar.pdf}}{\includegraphics[width=0.49 \textwidth]{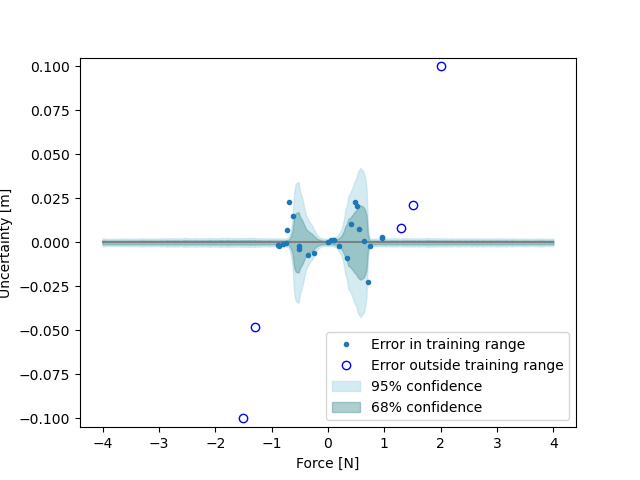}\label{Beam_MLE}}} 
     \subfloat[]{\includegraphics[width=0.49 \textwidth]{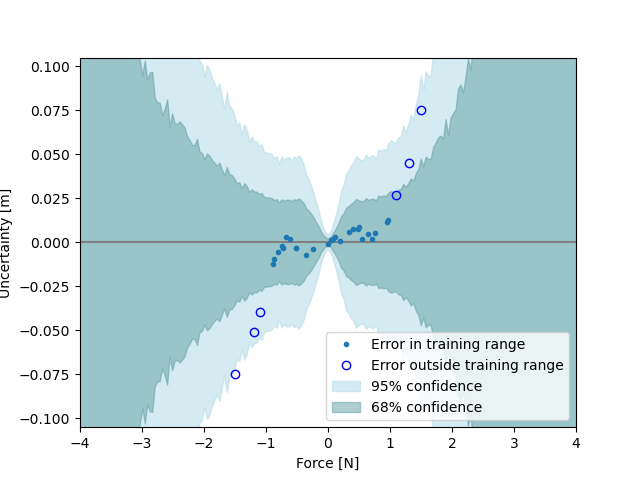}\label{Beam_VB}}
     
     \subfloat[]{\stackinset{r}{.31\textwidth}{t}{.055\textwidth}{\includegraphics[width=0.10\textwidth]{Images/inset_L.pdf}}{\includegraphics[width=0.49 \textwidth]{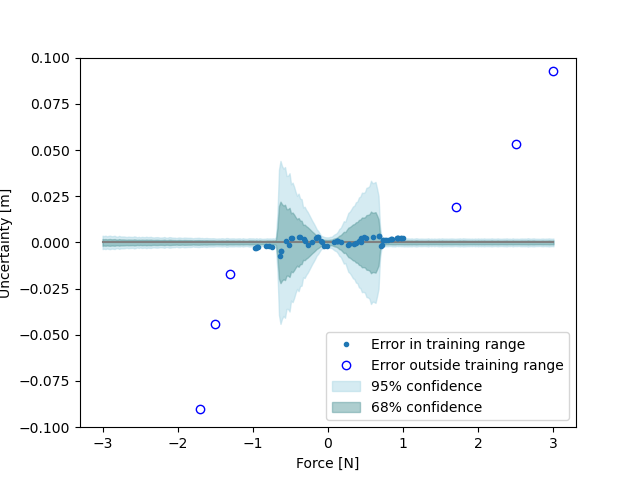}\label{L_MLE}}}
     \subfloat[]{\includegraphics[width=0.49 \textwidth]{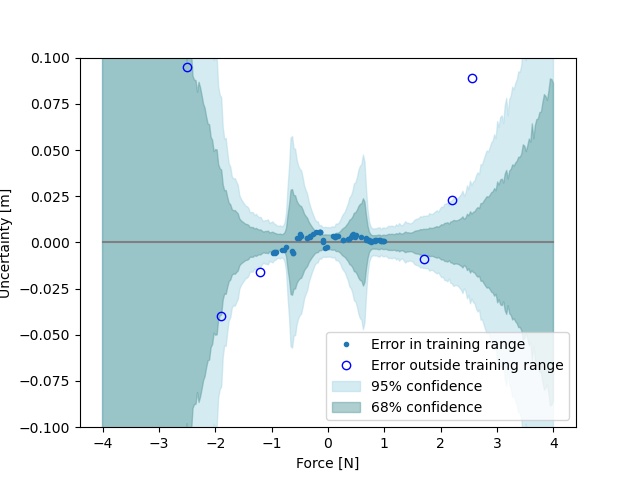}\label{L_VB}}

      \caption{Uncertainty predictions for noisy data cases using probabilistic U-Nets. For the 2D beam case (upper row), prediction is done for the Y-displacement of the corner node. For 2D L-shape (lower row), prediction is done for the X-displacement of the inner corner node (see insets). (a)\&(c) Uncertainty predictions using MLE approach. MLE fails to capture the uncertainty outside the training region. (b)\&(d) Uncertainty predictions using Variational Bayes approach. VB is capable to capture the uncertainty outside the training region.}
     \label{fig: MLEvsBayes}
\end{figure}

For both 2D beam and 2D L-shape cases, we modify the existing datasets (of the input force range -1 N to 1 N as shown Table~\ref{tab:datasets}) by incorporating random noises to displacement values. When the magnitude of applied force is less than 0.7 N, we add a random noise (from a continuous uniform distribution) within 20\% of the real displacement solutions, i.e., when $\norm{\bm{f}}_2\leqslant 0.7$ we set $\bm{u}\rightarrow (1+\gamma)\bm{u}$, where $\gamma \sim \text{U}(-0.2,0.2)$. Now, the probabilistic networks (MLE and Variational Bayes) are trained using these noisy datasets. In the prediction phase, we apply forces to a single chosen corner node in a single direction, with magnitudes ranging from -4 N to 4 N (see insets in Fig.~\ref{fig: MLEvsBayes}). Then we analyse the predicted uncertainties associated with displacements of respective nodes to which the force has been applied, and how they relate to the level of input force noises.

Figures~\ref{Beam_MLE} and \ref{L_MLE} show that the MLE approach is able to capture the noises in the training data region, although it fails to produce reliable uncertainty estimates outside that region (extrapolated region). The network is very confident in predictions even though we move away from the training region, and the prediction errors there are clearly visible. Whereas from Figure~\ref{Beam_VB}-\ref{L_VB}, we can see that the Variational Bayes approach is able to capture both effects: the effect of noises in the data, as well as the desired effect of gradually increasing uncertainty as we move away from the training region. 

\subsection{Prediction and training times}\label{sec: training and pred times}

\textbf{Prediction Times}

Although networks are trained on Graphical Processing Units (GPU), predictions are computationally inexpensive on user end Central Processing Units (CPU) as well. Also, since recent years, GPU cloud computing is easily accessible, one can leverage GPU support over the internet. All these factors make our framework easily deployable to the user end. Table~\ref{Tab: prediction_times} gives the comparison of prediction times for different examples on GPU as well on CPU.

\begin{table}[h]
\begin{center}
 \begin{tabular}{c | c | c | c | c | c | c } 
 Type & dof & $t\_\text{fem}_{\text{CPU}}$~[s] & $t_{\text{CPU}}$~[s] & $t_{\text{GPU}}$~[s] & $\frac{t\_\text{fem}_{\text{CPU}}}{t_{\text{CPU}}}$  & $\frac{t\_fem_{\text{CPU}}}{t_{\text{GPU}}}$\\ [0.5ex] 
 \hline 
2D Beam & 128 & 0.123 & 0.005 & 0.001 & 25 & 123 \\
2D L-shape & 256 &  0.120 & 0.007 &0.001 &17 & 120\\ 
3D Beam & 12096 & 3.1  &0.1 &  0.009 & 31 & 345 
\end{tabular}
\end{center}
\caption{Prediction times of deterministic U-Net on CPU, GPU. Under similar computational resources on CPU, U-Net shows 31 times speedup, which can be even more depending on the boundary conditions. Whereas GPU shows nearly 350 times speedup.}
\label{Tab: prediction_times}
\end{table}

For some of the force values in the testing set, the FEM solution took more than $3$s. Hence under identical computational resources, deterministic U-Net gave 31 times speedup. Another important point to mention here is, both deterministic and Bayesian U-Net, individually take the same time for prediction irrespective of the value of the input (applied force). In the case of the FEM, solution time evolves with the value of applied force. This is because we use an iterative solver and adaptive load-stepping scheme to avoid convergence issues for large load cases. Hence on local, we can expect much more speedup than 31 times when we go towards the higher input force values. Deterministic U-Nets gave nearly 350 times speed up when predictions were done using GPU. Even with the high dimensional 3D examples, U-Net did not take more than 10 ms, thus satisfying the real-time constraint. \vspace{2mm}

The time of prediction of Bayesian U-Net is the time of sampling over output distribution, which is as long as 300 stochastic forward passes in our case. For the above example (for both 2D beam and 2D L-shape), 300 forward passes for a single test example took 0.1 secs. Compared to the deterministic case, the average time for the prediction of single-pass is less (0.3 ms) because of the efficient utilisation of batch prediction. Hence even Bayesian inference takes very little time in the prediction phase.

\textbf{Training Times}

For any neural network, the training phase of the model is the most resource-intensive task. Hence modern machine learning open source libraries such as Tensorflow, Keras, PyTorch are optimized to work with GPUs. GPU has a parallel structure that offers faster computing and increased efficiency compared to the user end computer with its CPU. Table~\ref{Tab: training_times} gives GPU training times for different datasets for both deterministic and probabilistic U-Nets.

\begin{table}[h]
\begin{center}
 \begin{tabular}{l | c |c | c } 
 Example & Dataset size, $N$ & $t_{\text{train}}~[\text{min}]$ & N. of trainable parameters \\[0.5ex] 
 \hline
2D Beam (D)& 5700 & 131 & 7.5 E+6  \\
2D Beam (VB)& 5700 & 226 & 15.1 E+6 \\
\hline
2D L-Shaped (D)& 3800 & 78 & 7.5 E+6 \\
2D L-Shaped (VB) & 3800 & 143 & 14.6 E+6  \\
\hline
3D Beam (D) & 33688 & 1060 & 94.1 E+6 

\end{tabular}
\end{center}
\caption{U-Net training times, $t_{\text{train}}$. D = Deterministic, VB = Variational Bayes.}
\label{Tab: training_times}
\end{table}

Bayesian U-Nets have more parameters to be trained, additionally, we need to sample over the approximate posterior as described in Section~\ref{sec: loss_VB}. Hence training times for Bayesian U-Nets are significantly higher than for the deterministic counterparts. As the size of the problem grows, training time proportionally increases as well. Hence the training time for the 3D beam case is higher compared to 2D cases. Note however, that this time can be reduced by opting alternate topologies of U-Nets, and one way of doing so is keeping a constant number of channels in each U-Net level instead of increasing it (which will be analyzed below).
%\vspace{2mm}

\textbf{Effect of number of channels}

The training time of U-Net can be reduced by decreasing the number of trainable parameters of the model, and one of the ways to achieve this is to decrease the number of channels at each U-Net level. This can have, however, a side effect on prediction accuracy (intuitively, channels are partially responsible for capturing nonlinearities). We analyze these competing effects by performing a case study for the deterministic 3D-Beam case for architectures with different constant (not variable) number of channels, $c$. 

Table~\ref{const_channel} shows a comparison of training times and prediction errors. As we can see, as compared to the architecture used in Section~\ref{sec: Deterministic}, the use of 64 channels at each level gave comparable error values, while the training time is about three times lower. We can also observe that an excessive increase in the number of channels ($c=128$) results in deterioration of not only training time but also the prediction accuracy, which can be interpreted as a well-known effect of overfitting. For reference, we have provided GPU prediction times for these networks as well.   

\begin{table}[h]
\begin{center}
 \begin{tabular}{c | c | c | c | c } 
 N. of channels, $c$ & $\Bar{e}$ [m] & $\sigma(e)~[\text{m}]$ & $t_{\text{train}}~[\text{min}]$ & $t_{\text{GPU}}$~[ms] \\ [0.5ex] 
 \hline
 16 & 1.6 E-3 & 0.9 E-3  & 272 & 6\\ 
 32 & 1.1 E-3 & 0.7 E-3  & 293 & 6.5 \\
 \textbf{64} & \textbf{0.8 E-3} & \textbf{0.5 E-3}  & \textbf{348} & 7.5 \\
 128 & 3.5 E-3 & 3.3 E-3  & 646 & 9  \\
\end{tabular}
\end{center}
\caption{Error metrics, and training and predicting times for different deterministic U-Net architectures trained on the 3D dataset. A constant number of channels, $c$, is used in each level of U-Net. The use of 64 channels is optimum to achieve error and computational time trade-off. Standard 3D U-Net architecture took 1060 mins to train on the identical dataset.}
\label{const_channel}
\end{table}

\section{Conclusions}\label{sec: Conclusion}

In this work, we have proposed a deterministic/probabilistic neural network framework that is capable of accurately predicting large deformations in real-time. Although in the present work we only used artificially generated data for training, the framework can naturally assimilate experimental data as well. Because of these factors, our framework has the potential for data-driven applications requiring very fast response rates, such as patient-specific computer-aided surgery of soft human tissues. 

In addition to the predictions, the proposed probabilistic framework is also capable of giving reliable uncertainty estimates. Indeed, we showed that the predicted uncertainties correlate with the prediction errors (fitting errors to FEM solution). We also showed that the uncertainties rapidly increase in the extrapolated region, which is the desired property that we expected to achieve. Additionally, we were able to capture the noises present in the data, which has been validated with two probabilistic approaches (Maximum Likelihood Estimation and Variational Bayes). As such, our framework can be seen as a step towards making real-time large-deformation simulations more trustworthy.

To the best of our knowledge, this is the first time the state-of-the-art Bayesian Neural Networks are used in the context of non-linear body deformations. We believe that this work can serve as a reference for further developments in this emerging area of research. Due to its potentially high efficiency and accuracy, as well as due to its unique probabilistic predictive capabilities, we believe that the presented framework will turn out to be useful in a wide scope of novel engineering applications. 

Besides showing promising results, we also demonstrated several important limitations of the current framework. Firstly, the convolution operations that are used in our U-Nets' implementation require structured meshes. We showed in the paper possible methods to extend our framework to unstructured meshes, which can be done with a moderate effort in the future. Secondly, we observed that the proposed novel technique to quantify uncertainties in extrapolated regions does not always give reliable predictions. Bayesian U-Nets failed to give reliable credible intervals of predictions when we applied the force on the nodes which were not part of the training procedure (i.e. extrapolated data in the spatial dimensions). As discussed in the paper, it seems to be a more fundamental and challenging problem that needs a dedicated approach, which is left for future research.

\vspace{5mm}

\emph{Acknowledgements:}\\

\begin{wrapfigure}{l}{0.34\textwidth}
    \includegraphics[width=0.3\textwidth]{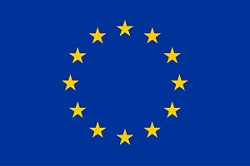}
\end{wrapfigure}

This project has received funding from the European Union’s Horizon 2020 research and innovation programme under the Marie Sklodowska-Curie grant agreement No. 764644. 
Jakub Lengiewicz would like to acknowledge the support from EU Horizon 2020 Marie Sklodowska Curie Individual Fellowship \emph{MOrPhEM} under Grant 800150.
This paper only contains the author's views and the Research Executive Agency and the Commission are not responsible for any use that may be made of the information it contains. 

Stephane Bordas and Jakub Lengiewicz are grateful for the support of the Fonds National de la Recherche Luxembourg FNR grant QuaC C20/MS/14782078. Stephane Bordas received funding from the European Union's Horizon 2020 research and innovation programme under grant agreement No 811099 TWINNING Project DRIVEN for the University
of Luxembourg.

\bibliography{mybibfile}

\end{document}